\documentclass[letterpaper, 10 pt, conference]{ieeeconf}  % Comment this line out if you need a4paper

\IEEEoverridecommandlockouts                              % This command is only needed if
\usepackage{graphicx} % for pdf, bitmapped graphics files
\usepackage{cite} % smarter citations, e.g. makes [1][2][3] into [1]-[3]
\usepackage[caption=false,font=scriptsize,justification=raggedright,singlelinecheck=true]{subfig} % subfigures
\usepackage{verbatim}  % multiline comments via \begin{comment}\end{comment}
\usepackage{url} % \url{my_url_here}.
\usepackage[multidot]{grffile} % allow dots in graphics filenames
\usepackage{times} % assumes new font selection scheme installed
\usepackage{amsmath} % assumes amsmath package installed
\usepackage{amssymb}  % assumes amsmath package installed
\usepackage[usenames,dvipsnames]{xcolor}
\usepackage[ colorlinks=true, citecolor=black, linkcolor=black, urlcolor=black,  bookmarks=true, bookmarksnumbered=true, pdftex, plainpages=false, pdfpagelabels]{hyperref}  % hyperref no worky
\usepackage{color}
\usepackage{ctable}

\usepackage{booktabs}
\usepackage{array}
\usepackage{longtable}
\newcolumntype{M}[1]{>{\centering\arraybackslash}m{#1}}
\newcolumntype{A}{>{\begin{minipage}[c][3cm][c]{2cm}\centering\arraybackslash}m{2cm}<{\end{minipage}}}
\usepackage{xifthen}

\newcommand{\bi}{\begin{itemize}}
\newcommand{\ei}{\end{itemize}}

\newcommand{\bfig}{\begin{figure}}
\newcommand{\efig}{\end{figure}}

\newcommand{\be}{\begin{equation}}
\newcommand{\ee}{\end{equation}}

\newcommand{\ba}{\begin{eqnarray}}
\newcommand{\ea}{\end{eqnarray}}

\newcommand{\eq}[1]{Eq.~\ref{#1}}

\newcommand{\REF}[2][ZZZZ]{\ifthenelse{\equal{#1}{ZZZZ}}% true, no optional arg
  {\index{general}{#2}\ifthenelse{\boolean{draft}}{{\color{red}\it#2}}{#2}}% false, optional arg
  {\index{general}{#1}\ifthenelse{\boolean{draft}}{{\color{red}\it#2}}{#2}}}

\newcommand{\DEF}[2][ZZZZ]{\ifthenelse{\equal{#1}{ZZZZ}}% true, no optional arg
  {\index{general}{#2}\ifthenelse{\boolean{draft}}{{\color{red}\it#2}}{#2}}% false, optional arg
  {\index{general}{#1}\ifthenelse{\boolean{draft}}{{\color{red}\it#2}}{#2}}}

\newcommand{\DEFX}[2][ZZZZ]{\ifthenelse{\equal{#1}{ZZZZ}}% true, no optional arg
  {\index{general}{#2|textbf}\ifthenelse{\boolean{draft}}{{\color{red}\it#2}}{#2}}% false, optional arg
  {\index{general}{#1|textbf}\ifthenelse{\boolean{draft}}{{\color{red}\it#2}}{#2}}}

\newcommand{\model}[1]{\index{code}{#1@\textit{#1}}\ifthenelse{\boolean{draft}}{{\color{green}\Verb+#1+}}{\Verb+#1+}}
\newcommand{\block}[1]{\ifthenelse{\boolean{draft}}{{\color{green}\Verb+#1+}}{\textsf{#1}}}

\newcommand{\func}[2][ZZZZ]{\ifthenelse{\equal{#1}{ZZZZ}}{\index{code}{#2}}{\index{code}{#1}}\ifthenelse{\boolean{draft}}{{\color{green}\Verb+#2+}}{\Verb+#2+}}

\newcommand{\methodb}[2]{\index{code}{#1@\textbf{#1}!.#2}\ifthenelse{\boolean{draft}}{{\color{magenta}\Verb+#1.#2+}}{\Verb+#1.#2+}}
\newcommand{\method}[2]{\index{code}{#1@\textbf{#1}!.#2}\ifthenelse{\boolean{draft}}{{\color{magenta}\Verb+#2+}}{\Verb+#2+}}
\newcommand{\class}[1]{\index{code}{#1@\textbf{#1}}\ifthenelse{\boolean{draft}}{{\color{cyan}\Verb+#1+}}{\Verb+#1+}}
\newcommand{\property}[1]{\index{property}{#1}\ifthenelse{\boolean{draft}}{{\color{cyan}\Verb+#1+}}{\Verb+#1+}}

\newcommand{\presup}[1]{\,{}^{\scriptscriptstyle #1}\!}

\newcommand{\pose}[1][ZZZZ]{\ifthenelse{\equal{#1}{ZZZZ}}{}{\presup{#1}}{\mathbf{\xi}}}

\usepackage{mathrsfs}

\newcommand{\poser}[1][ZZZZ]{\ifthenelse{\equal{#1}{ZZZZ}}{}{\presup{#1}}{\mathscr{R}}}
\newcommand{\poserx}[1][ZZZZ]{\ifthenelse{\equal{#1}{ZZZZ}}{}{\presup{#1}}{\mathscr{R}_x}}
\newcommand{\posery}[1][ZZZZ]{\ifthenelse{\equal{#1}{ZZZZ}}{}{\presup{#1}}{\mathscr{R}_y}}
\newcommand{\poserz}[1][ZZZZ]{\ifthenelse{\equal{#1}{ZZZZ}}{}{\presup{#1}}{\mathscr{R}_z}}
\newcommand{\poserw}[1][ZZZZ]{\ifthenelse{\equal{#1}{ZZZZ}}{}{\presup{#1}}{\mathscr{R}_\omega}}
\newcommand{\poset}[1][ZZZZ]{\ifthenelse{\equal{#1}{ZZZZ}}{}{\presup{#1}}{\mathscr{T}}}
\newcommand{\posetx}[1][ZZZZ]{\ifthenelse{\equal{#1}{ZZZZ}}{}{\presup{#1}}{\mathscr{T}_x}}
\newcommand{\posety}[1][ZZZZ]{\ifthenelse{\equal{#1}{ZZZZ}}{}{\presup{#1}}{\mathscr{T}_y}}
\newcommand{\posetz}[1][ZZZZ]{\ifthenelse{\equal{#1}{ZZZZ}}{}{\presup{#1}}{\mathscr{T}_z}}
\newcommand{\posett}[1][ZZZZ]{\ifthenelse{\equal{#1}{ZZZZ}}{}{\presup{#1}}{\mathscr{T}\!}}

\newcommand{\poseri}[1][ZZZZ]{\ifthenelse{\equal{#1}{ZZZZ}}{}{\presup{#1}}{\mathscr{R}_i}}
\newcommand{\poseti}[1][ZZZZ]{\ifthenelse{\equal{#1}{ZZZZ}}{}{\presup{#1}}{\mathscr{T}_i}}

\newcommand{\twist}[1][ZZZZ]{\ifthenelse{\equal{#1}{ZZZZ}}{}{\presup{#1}}{S}}

\newcommand{\estpose}[1][ZZZZ]{\ifthenelse{\equal{#1}{ZZZZ}}{}{\presup{#1}}{\mathbf{\hat{\xi}}}}
\newcommand{\hpose}[1][ZZZZ]{\ifthenelse{\equal{#1}{ZZZZ}}{}{\presup{#1}}{\hat{\mathbf{\xi}}}}
\newcommand{\posedot}[1][ZZZZ]{\ifthenelse{\equal{#1}{ZZZZ}}{}{\presup{#1}}{\mathbf{\nu}}}
\newcommand{\q}[1][ZZZZ]{\ifthenelse{\equal{#1}{ZZZZ}}{}{\presup{#1}}{\mathring{q}}}

\DeclareMathAlphabet{\mathitbf}{OML}{cmm}{b}{it}
\renewcommand{\vec}[2][ZZZZ]{\ifthenelse{\equal{#1}{ZZZZ}}{}{\presup{#1}}{\mathitbf{#2}}}

\newcommand{\hvec}[2][ZZZZ]{\ifthenelse{\equal{#1}{ZZZZ}}{}{\presup{#1}}{\hat{\vec{#2}}}}
\newcommand{\ovec}[2][ZZZZ]{\ifthenelse{\equal{#1}{ZZZZ}}{}{\presup{#1}}{\mathring{\vec{#2}}}}
\newcommand{\tvec}[2][ZZZZ]{\ifthenelse{\equal{#1}{ZZZZ}}{}{\presup{#1}}{\tilde{\vec{#2}}}}
\newcommand{\evec}[2][ZZZZ]{\ifthenelse{\equal{#1}{ZZZZ}}{}{\presup{#1}}{\hat{\vec{#2}}}}
\newcommand{\dvec}[2][ZZZZ]{\ifthenelse{\equal{#1}{ZZZZ}}{}{\presup{#1}}{\dot{\vec{#2}}}}
\newcommand{\ddvec}[2][ZZZZ]{\ifthenelse{\equal{#1}{ZZZZ}}{}{\presup{#1}}{\ddot{\vec{#2}}}}
\newcommand{\vech}[2][ZZZZ]{\ifthenelse{\equal{#1}{ZZZZ}}{}{\presup{#1}}{\mathitbf{\tilde{#2}}}}
\newcommand{\vecb}[2][ZZZZ]{\ifthenelse{\equal{#1}{ZZZZ}}{}{\presup{#1}}{\bar{\underline #2}}}
\newcommand{\mat}[2][ZZZZ]{\ifthenelse{\equal{#1}{ZZZZ}}{}{\presup{#1}\,}{{\boldsymbol #2}}}
\newcommand{\hmat}[2][ZZZZ]{\ifthenelse{\equal{#1}{ZZZZ}}{}{\presup{#1}\,}{{\hat{\boldsymbol #2}}}}
\newcommand{\dmat}[2][ZZZZ]{\ifthenelse{\equal{#1}{ZZZZ}}{}{\presup{#1}\,}{\dot{\boldsymbol #2}}}
\newcommand{\emat}[2][ZZZZ]{\ifthenelse{\equal{#1}{ZZZZ}}{}{\presup{#1}\,}{\hat{\boldsymbol #2}}}
\newcommand{\matfn}[3][ZZZZ]{\ifthenelse{\equal{#1}{ZZZZ}}{}{\presup{#1}}{{\mat{#2}}\left(#3\right)}}
\newcommand{\Rt}[2][ZZZZ]{\ifthenelse{\equal{#1}{ZZZZ}}{}{\presup{#1}}{{\bf R}\left(#2\right)}}

\newcommand{\point}[2][ZZZZ]{\ifthenelse{\equal{#1}{ZZZZ}}{}{\presup{#1}}{\mathbf{\mathrm{#2}}}}
\renewcommand{\frame}[3][ZZZZ]{\ifthenelse{\equal{#1}{ZZZZ}}{}{\presup{#1}}{\mat{#2}}_{#3}}
\newcommand{\frameh}[3][ZZZZ]{\ifthenelse{\equal{#1}{ZZZZ}}{}{\presup{#1}}{\hat{#2}}_{#3}}
\newcommand{\frameb}[3][ZZZZ]{\ifthenelse{\equal{#1}{ZZZZ}}{}{\presup{#1}}{\bar{#2}}_{#3}}

\newcommand{\pnt}[2][ZZZZ]{\ifthenelse{\equal{#1}{ZZZZ}}{}{\presup{#1}}{\mathbf{#2}}}
\newfont{\School}{pncr}
\newfont{\eightTR}{pncr at 8pt}
\newcommand{\dorian}[1]{{\color{ForestGreen}dt: #1}}

\title{\LARGE \bf
Distinguishing Refracted Features using Light Field Cameras with Application to Structure from Motion
}

\author{Dorian Tsai$^{1}$, Donald G. Dansereau$^{2}$, Thierry Peynot$^{1}$ and Peter Corke$^{1}$% <-this % stops a space
\thanks{This research was partly supported by the Australian Research Council (ARC) Centre of Excellence for Robotic Vision (CE140100016).
}
\thanks{$^{1}$D. Tsai, T. Peynot and  P. Corke are with the Australian Centre for Robotic Vision (ACRV), Queensland University of Technology (QUT), Brisbane, Australia {\tt\small \{dy.tsai, t.peynot, peter.corke\}@qut.edu.au}}
\thanks{$^{2}$D. Dansereau is with the Stanford Computational Imaging Lab, Stanford University, CA, USA.  {\tt\small donald.dansereau@gmail.com}}
}

\begin{document}

\maketitle
\thispagestyle{empty}
\pagestyle{empty}

%%%%%%%%%%%%%%%%%%%%%%%%%%%%%%%%%%%%%%%%%%%%%%%%%%%%%%%%%%%%%%%%%%%%%%%%%%%%%%%%
\begin{abstract}

Robots must reliably interact with refractive objects in many applications; however, refractive objects can cause many robotic vision algorithms to become unreliable or even fail, particularly feature-based matching applications, such as structure-from-motion.
% visual odometry, visual servoing, and simultaneous localization and mapping (SLAM).
We propose a method to distinguish between refracted and Lambertian image features using a light field camera. Specifically, we propose to use textural cross-correlation to characterise apparent feature motion in a single light field, and compare this motion to its Lambertian equivalent based on 4D light field geometry.
%Refractive objects cause problems in robotic vision, particularly in feature-based structure-from-motion. % ultimately, we want to recreate the objects themselves
% We develop a method to distinguish between refractive and Lambertian  image features using the light field.
Our refracted feature distinguisher has a 34.3\% higher rate of detection compared to state-of-the-art for light fields captured with large baselines relative to the refractive object. Our method also applies to light field cameras with much smaller baselines than previously considered, yielding up to 2 times better detection for 2D-refractive objects, such as a sphere, and up to 8 times better for 1D-refractive objects, such as a cylinder. % how to generalize the LF cylinder case?
% We also extend our method to light field cameras with much smaller baselines than previously considered.
For structure from motion, we demonstrate that rejecting refracted features using our distinguisher yields up to 42.4\% lower reprojection error, and lower failure rate when the robot is approaching refractive objects.
% We develop a light field feature detector that is sensitive to non-Lambertian surfaces, such as refractive objects. Our detector out-performs the state-of-the-art in terms of (some specific performance metric) for structure-from-motion in a scene containing refractive objects.
Our method lead to more robust robot vision in the presence of refractive objects.
% Really, we are developing a LF descriptor, since we are not changing where we find the location of the feature (just Tosic).

\end{abstract}

%%%%%%%%%%%%%%%%%%%%%%%%%%%%%%%%%%%%%%%%%%%%%%%%%%%%%%%%%%%%%%%%%%%%%%%%%%%%%%%%
\section{INTRODUCTION}

% Provide a brief motivation for the article (transparent objects, light fields, light field features). Address non-LF-based approaches, and their limitations.
% Outline of paper. Main contributions.

\begin{comment}
\begin{itemize}
  \item problem/motivation case w refractive objects, intro the context of SfM/moving robot towards refractive objects
  \item consider L:F as a possible solution
  \item list our contributions as bulleted list (see email)
  \item brief paper outline
  \item summary figure, Lytro Illum image w refractive object (cylinder) identified along with colour image for comparison, consider showing curved EPIs/extracted curves of refractive feature
\end{itemize}
\end{comment}

Robots for the real world will inevitably interact with refractive objects.
Robots must contend with wine glasses and clear water bottles in domestic applications~\cite{kemp2007challengesRobotManipulation}; glass and clear plastic packaging for quality assessment and packing in manufacturing~\cite{ihrke2010transparentSurvey}; as well as water and ice for outdoor operations~\cite{dansereau2014Thesis}. All of these applications typically require either object structure and/or robot motion to automate.
Structure from motion (SfM) is a technique to recover both scene structure and camera pose from 2D images, and is widely applicable to many systems in computer and robotic vision~\cite{hartley2003multiViewGeometry,wei2013structureFromMotionSurvey}.
Many of these systems assume the scene is Lambertian, in that a 3D point's appearance in an image does not change significantly with viewpoint. However, non-Lambertian effects, including specular reflections, occlusions, and refraction, violate this assumption.
% should probably find a better example of challenging underwater robots or ice robots - Anne Jordt (Refractive SfM on Underwater Images?)
% Medical robotics must also contend with transparent objects, such as eyes and glass or plastic tools~\cite{queiroz2014specularEndoscope}.
% Clearly, refractive objects are relevant; however,
They pose a major problem for modern robotic vision systems because their appearance  depends on the camera's viewing pose and the visual texture of the object's background.

\begin{figure}[t]
\centering
\subfloat{\includegraphics[width=0.49\columnwidth]{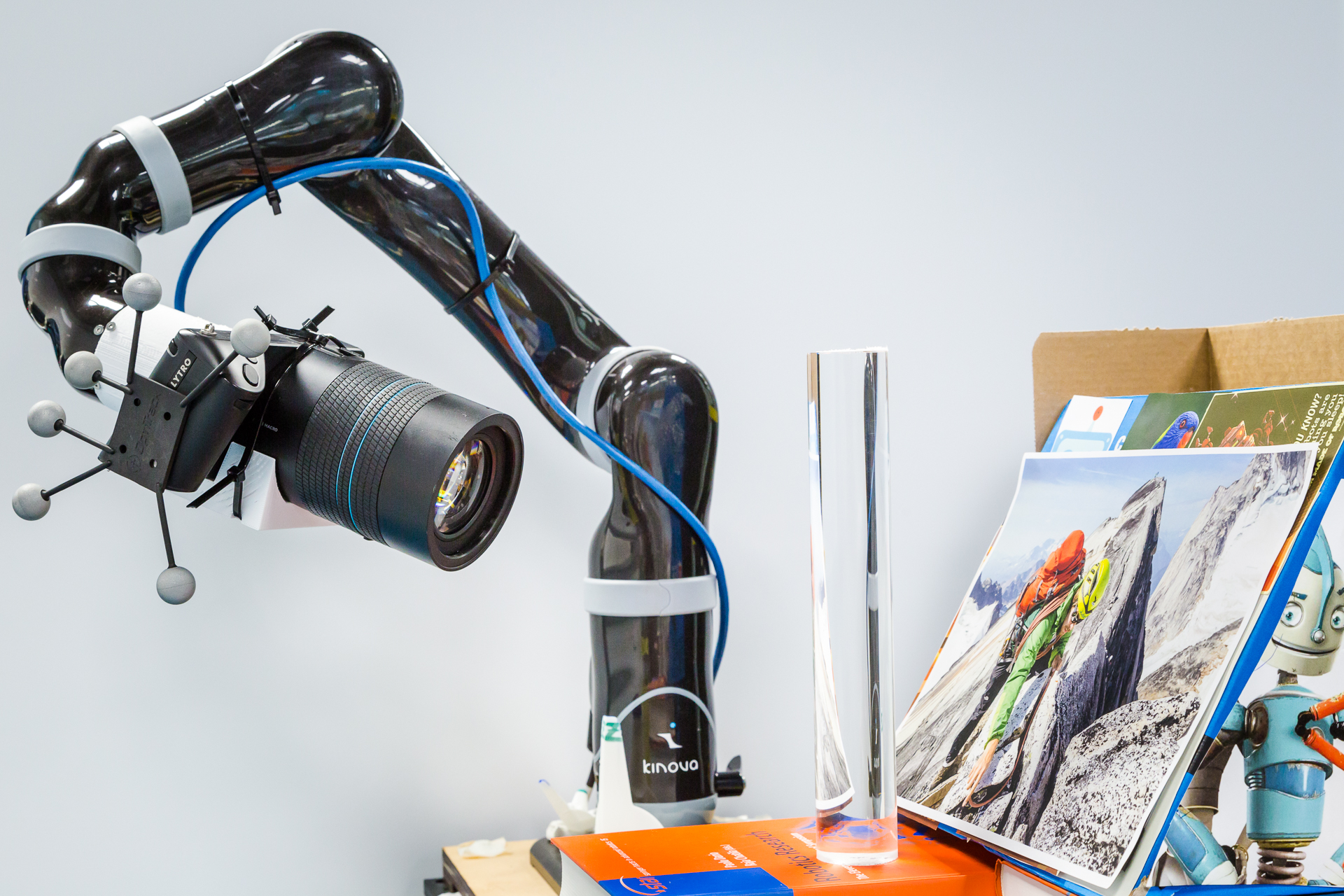}
\label{fig:title2}}
\subfloat{\includegraphics[width=0.49\columnwidth]{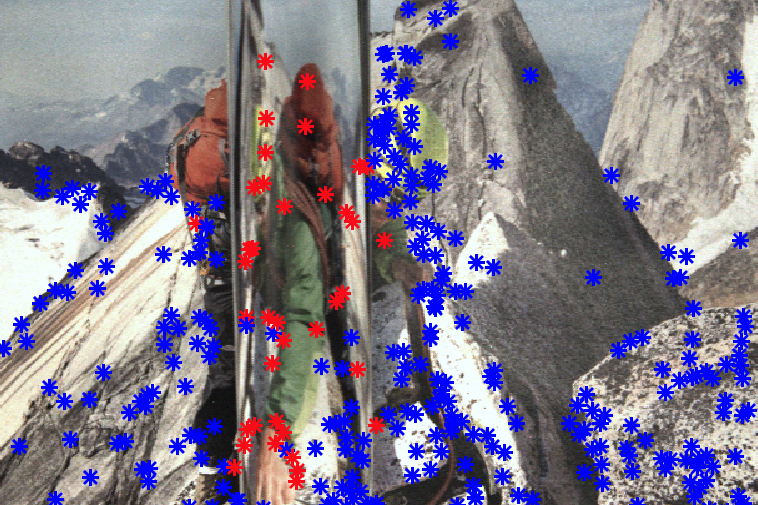}
\label{fig:titleSub}}\\
\caption{(left) A light field camera mounted on a robot arm was used to distinguish refractive objects in a scene. (right) SIFT features that were  distinguished as Lambertian (blue) and refracted (red), indicating the refractive cylinder in the middle of the scene.}
\label{fig:title}
\vspace{-1.5em}
\end{figure}

Image features are distinct points of interest in the scene that can be repeatedly and reliably identified from different viewpoints, and have been used in SfM, but also many other robotic vision algorithms, such as object recognition, image segmentation, visual servoing, visual odometry, and simultaneous localization and mapping (SLAM). In SfM, features are often used for image registration. When reconstructing a scene containing a refractive object, such as Fig.~\ref{fig:title}, image features occluded by the object appear to move differently from the rest of the scene. They can cause inconsistencies, errors, and even failures for modern robotic vision systems.
% need strong citation example?

Light field (LF) cameras offer a potential solution to the problem of refractive objects. LF cameras simultaneously capture multiple images of the same scene from different viewpoints in a regular and dense sampling.
% A useful way of thinking about LF cameras is as an array of monocular cameras, each providing a view of the same scene from slightly different viewpoints, and imaged simultaneously to provide a 2D tiling of 2D images, which we know as the light field.
% Light field motivation: Most modern RV systems predominantly rely on monocular cameras cannot distinguish transparent objects without known motion (not ideal because solutions rely on complex/iterative/ill-posed methods), or different/known lighting (not practical for mobile robotic applications).
% Stereo and trinocular camera systems do not provide sufficient detail, since they also usually rely on similar appearances for correspondence matching. Mention other non-LF methods for recovering the features on/of transparent objects, but they are not practical for mobile robot platforms for other reasons.
The LF could allow robots to more reliably and efficiently capture the behaviour of refractive objects in a single shot by exploiting the known geometry of the multiple views.
% As a first step towards dealing with refractive objects,
We take 2D image features from the central view of the LF, and determine which of these exhibit refractive behaviour in the 4D LF, which we refer to as a refracted feature, and use this as a method of distinguishing good features for SfM.

% define features
% One of the challenges with light field imaging in the context of SfM is dealing with the large amount of redundant image data, and its associated high bandwidth when capturing multiple light fields from different positions in the world. Features are unique and distinguishable points in the world that abstract image data to help correspond two light fields. Features have been used throughout computer and robotic vision, and often serve as a basis for many vision-based algorithms.
% However, the topic of LF features, particularly in the area of transparent objects, has not yet been well explored.
% \newpage % just to ensure contributions kept together (mbox?)

Our main contributions are the following.
\begin{itemize}
% \item \don{We draw inspiration from, and draw on the previous work from...}
\item We extend previous work to develop a light field feature distinguisher for refractive objects. In particular, we detect the differences between the apparent motion of non-Lambertian and Lambertian features in the 4D light field to distinguish refractive objects more reliably than previous work.
\item We propose a novel approach to describe the apparent motion of a feature observed within the 4D light field based on textural cross-correlation.
% \item We provide a publicly available implementation of LF features.
\item We extend refracted feature distinguishing capabilities to lenslet-based LF cameras that are limited to much smaller baselines by considering non-uniform, non-Lambertian apparent motion in the light field. All light fields captured for these experiments will be available online at \textit{https://tinyurl.com/LFRefractive}.
\end{itemize}

Our method outperforms the state of the art in terms of detecting refracted features, including small-baseline LF cameras for the first time. We also show that rejecting refracted features before the SfM pipeline can yield lower reprojection error in the presence of refractive objects, provided there are a sufficient number of Lambertian features remaining.
% \item \don{We also show significantly denser reconstructions in the presence of refractive objects using conventional SfM with our novel LF features.}

% \dorian{consider context of visual servoing?}
The main limitation of our method is that it requires  background visual texture distorted by the refractive object. Our method's effectiveness depends on the extent to which the appearance of the object is warped in the light field. This depends on the geometry and refractive indices of the object involved.
% One of the main limitations of this work is the reliance on visual texture and the autocorrelation function. Our method cannot handle refractive objects when there is no background texture.
% Additionally for lenslet-based cameras, our method does not distinguish refractive features when the apparent motion is consistent in different viewing directions~\dorian{which occurs when?}.
% when does this occur?
% Our method relies on thresholds that we have set empirically, which results in limited flexibility to changing environmental conditions.
% Ridge following applied to the feature curve would likely greatly improve the robustness of the extraction process.

The remainder of this paper is organized as follows. We describe the related work, provide background on LF geometry, and explain our method for distinguishing refracted features. Then we show our experimental results for extraction of a feature's apparent motion, detection with different camera baselines and object types, and validation in the context of monocular SfM. Finally, we conclude the paper and explore future work.
% The remainder of this paper is organized as follows. Section II describes the related work. Section III provides background on LF geometry. Section IV explains our method for distinguishing refractive features in the LF. Section V shows our experimental results for feature curve extraction, detection for different LF camera baselines and refractive object types to state-of-the-art, and illustrates the impact of rejecting refractive features in the context of monocular SfM. Lastly in Section VI, we conclude the paper and explore future work.

%%%%%%%%%%%%%%%%%%%%%%%%%%%%%%%%%%%%%%%%%%%%%%%%%%%%%%%%%%%%%%%%%%%%%%%%%%%%%%%%
\section{RELATED WORK}

% the purpose of this literature review is to carve out our niche - this is the landscape of how LF cameras deal with refractive objects, and in particular, how refractive objects are identified.

% what has been done with refractive objects in computer vision
A variety of strategies for detecting, and even reconstructing refractive objects using vision have been investigated~\cite{ihrke2010transparentSurvey}.
% ~\cite{morris2007inhomogeneousObjects,wetzstein2011refractive}.
% \dorian{List a few examples from confirmation report, and 1 sentence on how they work (Ben-Ezra, Ihrke, Morris)}.
% For example, reflectivity has been used to reconstruct refractive object shape. A single monocular camera with a moving light source in a square grid has been used to densely reconstruct complex refractive objects by tracing the specular reflections from different, known lighting positions over multiple monocular images~\cite{morris2007inhomogeneousObjects}.
%Additionally, light refracted by transparent objects tends to be polarized, and thus a rotating polarizer in front of a monocular camera has been used to reconstruct glass objects~\cite{Miyazaki2005polarization}, but required placing the object within a dome of light to ensure even reflections.
% Shape from refraction has also been obtained using a monocular camera, and placing the object in front of a special optical sheet and lighting system, known as a light field probe~\cite{wetzstein2011refractive}.
However, many of these methods require known light sources with bulky configurations that are impractical for mobile robot applications.
% Recent work has been aimed at finding refractive objects within a single monocular image.
% SIFT features and a learning-based approach have been used to detect refractive objects, and provide their location as a bounding box~\cite{fritz2009learningtransparentObjects}. However, this approach required many training images from a variety of refractive objects under different lighting environments and backgrounds.
Multiple monocular images have been used to recover refractive object shape and pose~\cite{benezra2003transparent}; however,
% it is assumed that the background is far away, and that the object can be described using a parametric model. Importantly,
image features were manually tagged throughout camera motion, emphasizing the difficulty of automatically identifying and tracking refracted features due to the severe magnification of the background, and changes in image intensities when passing through the object.

% issue: most of my previous research is focused on shape reconstruction of refractive objects. And I do not yet get object shape... I simply detect...

LFs have recently been used to explore refractive objects. Wanner et al. considered planar refractive surfaces and reconstructed different depth layers that accounted for both refraction through a thin sheet of glass, and the reflection caused by its glossy surface~\cite{wanner2013reflectiveTransparentFromEPI}. However, this work was limited to thin planar surfaces and single reflections. Which depth layer was Lambertian, reflective or refractive was not distinguished, and refractive objects that caused significant distortion were not handled. Although our work does not determine the dense structure of the refractive object, %and is aimed towards sparse feature distinction, 
our approach can distinguish features from objects that significantly distort the LF.
%  highly refractive objects, provided their behaviour is non-Lambertian in the 4D LF.
% object recognition - Maeno introduced the LFD idea for object recognition. This is done by characterizing EPI curves of refractive features (how curves actually obtained?), does not perform well for changes in viewpoint (as shown by their graphs), used a large camera baseline.
% Xu extends this idea to create a measure of hyperplane linearity (which we refer to as ``hyperplanarity''), used optical flow through the multiple views to track the features, done for transparent object segmentation, also used LF camera with very large baseline. We build on their idea by extending it to incorporate a planarity measure, as well as a slope consistency measure.

For refractive object recognition, Maeno et al. proposed a light field distortion feature (LFD), which models an object's refraction pattern as image distortion based on differences in the corresponding image points between the multiple views of the LF, captured by a large-baseline (relative to the refractive object) LF camera array~\cite{maeno2013light}. % consider showing a figure?
% The LFD does not rely on the appearance of the background.
%, and differentiates Lambertian and non-Lambertian image features.
However, the authors observed significantly poor recognition performance due to specular reflections, as well as changes in camera pose.
% so should this be a consideration in my experiments?
% And because their approach was designed for recognition, their method assumes only one refractive object per image, while our approach is aimed at a per-feature basis.

\begin{comment}
\begin{figure}
\centering
\subfloat[][]{\includegraphics[width=0.38\textwidth]{Figures/lambertianFeature.pdf}
\label{fig:XuLinear}}\\
\subfloat[][]{\includegraphics[width=0.38\textwidth]{Figures/refractiveFeature.pdf}
\label{fig:XuTransparent}}\\
\caption{(a) Projection of the linear behaviour of a Lambertian feature, and (b) the non-linear behaviour of a refracted feature with respect to linear motion along the viewpoints of a light field.}
\label{fig:XuFeatureBehaviour}
\vspace{-1.5em}
\end{figure}
\end{comment}

Xu et al. used the LFD as a basis for refractive object image segmentation~\cite{xu2015transparentObjectSegmentation}. Corresponding image features from all views in the LF were fitted to the single normal of a 4D hyperplane in the least-squares sense using singular value decomposition (SVD). The smallest singular value was taken as a measure of error to the hyperplane of best fit, for which a threshold was applied to distinguish refracted features.
% Xu also handled occlusions and put the initial segmentations into a graph segmentation framework to improve results.
However, as we will show in this paper, a 3D point cannot be described by a single hyperplane in 4D. Instead, it manifests as a plane in 4D that has two orthogonal normal vectors. Our approach builds on Xu's method, and solves for both normals to find the plane of best fit in 4D; thus allowing us to distinguish more types of refractive objects with a higher rate of detection.

Furthermore, a key difficulty in feature-based approaches in the LF is obtaining the corresponding feature locations between multiple views. Both Maeno and Xu used optical flow between two views for correspondence, which does not exploit the unique geometry of the LF. % While this may be fast and efficient for a  $5\times5$ LF camera array with a large baseline ($\sim5$cm), this approach does not scale well for LF plenoptic cameras that have more than $13\times13$ views with a much smaller baseline ($\sim2$mm).
We propose a novel textural cross-correlation method to correspond features in the LF by describing their apparent motion in the LF, which we refer to as feature curves. This method directly exploits LF geometry and provides insight on the 4D nature of features in the LF.

Our interest in LF cameras stems from robot applications that often have mass, power and size constraints. Thus, we are interested in employing compact lenslet-based LF cameras to deal with refractive objects. However, most previous works have utilized gantries~\cite{wanner2013reflectiveTransparentFromEPI}, or large camera arrays~\cite{maeno2013light,xu2015transparentObjectSegmentation}; their results do not transfer to LF cameras with much smaller baselines, where refractive behaviour is less apparent, as we show later. We then demonstrate the performance of our method over two different LF camera baselines, and two different LF camera architectures. We demonstrate refracted feature identification with a lenslet-based LF camera, which to the best of our knowledge, has not been done before.

% For image features, conventional features like SIFT~\cite{lowe2004distinctive}, and SURF~\cite{bay2008surf} have worked well for Lambertian scenes; however, they are limited to 2D images, and have been found to be unreliable for scenes containing refractive objects~\cite{kragic2002surveyVisualServo}.
For LF cameras, LF-specific features have been investigated.
SIFT features augmented with ``slope'', an LF-based property related to depth, were proposed by the authors for visual servoing using a LF camera~\cite{tsai2016lfvisualservo}; however, refraction was not considered in prior work.
% Ghasemi proposed a scale-invariant global feature descriptor based on a modified Hough transform~\cite{ghasemi2014scaleInvariantFeature}; however, we are interested in local features. More recently
Tosic developed a scale-invariant, single pixel, edge detector by finding local extrema in a combined scale, depth, and image space~\cite{tosic2014lightFieldScaleSpace}. However, these LF features did not differentiate between Lambertian and refracted features, nor were they designed for reliable matching between LFs captured from different viewpoints.
In this paper, we detect unique keypoints that reject refracted content and work well for SfM. % By pre-filtering for refractive features using our distinguisher, we mitigate many of the problems observed with SIFT around refractive objects.

Recent work by Teixeira found SIFT features in all views of the LF and projected them into their corresponding epipolar plane images (EPI). These projections were filtered and grouped onto straight lines in their respective EPIs, then counted. Features with higher counts were observed in more views, and thus considered more reliable~\cite{teixeira2017epipolarLightfieldSift}.
% A framework for comparing 4D to 2D image features in terms of repeatability is also proposed.
However, their approach
% only looks for SIFT features that consistently lie on lines in the EPIs and
% intentionally filters out any non-linear feature behaviour; thus, their approach
did not consider any non-linear feature behaviour, while our proposed method aims at detecting these non-Lambertian features, and is focused on characterising them. This could be useful for many feature-based algorithms, including recognition, segmentation, visual servoing, simultaneous localization and mapping, visual odometry, and SfM.
% \dorian{Not sure if they are looking for 3D lines in 3D space, or simply look for 2D lines in each layer of the EPI cube (ie, each EPI), but Eq. 1 in paper suggests the latter. Doing the former would be better, because you can more easily smooth across multiple views (in the same motion direction) and spatial dimensions. Ultimately, we want to do this in 4D!}

We are interested in exploring the impact of our refracted feature distinguisher in a SfM framework. While there has been significant development in SfM in recent years~\cite{wei2013structureFromMotionSurvey}, Johannsen was the first to consider LFs in the SfM framework~\cite{johannsen2015linear}. As a first step, our work does not yet explore LF-based SfM. Instead, we
% utilize Schoenberger's SfM implementation, known as Colmap~\cite{schoenberger2016sfmRevisited}, to
investigate SfM's performance with respect to refracted features, which has not yet been fully explored. We show that rejecting refracted features reduces reprojection error and failure rate near refractive objects.
% \dorian{finish on something like how our contribution/paper makes us unique in the field}

% \don{Building/combining work of Xu and Tosic, our detector is sensitive to refraction, and providing novel insight on 4D nature of feature geometry/LF.}

% \subsection{Light Field Structure-from-Motion}

%%%%%%%%%%%%%%%%%%%%%%%%%%%%%%%%%%%%%%%%%%%%%%%%%%%%%%%%%%%%%%%%%%%%%%%%%%%%%%%%
\section{LIGHT FIELD BACKGROUND}

We parameterize the LF using the relative two-plane parameterization (2PP)~\cite{tsai2016lfvisualservo}. A ray with coordinates $\vec{\phi}=[s,t,u,v]^T$, where $^T$ represents the vector transpose, is described by two points of intersection with two parallel reference planes; an $s,t$ plane conventionally closest to the camera, and a $u,v$ plane conventionally closer to the scene, separated by arbitrary distance $D$.
% In this parameterization, $u$ and $v$ are expressed relative to $s$ and $t$. If $D$ is set to the camera focal length, $u$ and $v$ can be are image coordinates, $s,t$ can be considered camera position or the viewpoint.
\begin{comment}
\begin{figure}[tpb]
    \centering
    \includegraphics[width=0.35\textwidth]{Figures/Lightfield-2PP-Dansereau.png}
    \caption{The two-plane parameterization of light rays. Point $P$ forms a ray $\phi$ that intersects the two parallel planes, separated by distance $D$~\cite{dansereau2014Thesis}.}
    \label{fig:loriGlassVertEpi}
\end{figure}
\end{comment}

% \subsection{A Point in 3D to a Plane in 4D}
For a Lambertian point in space $\vec{P} = [P_x, P_y, P_z]^T \in \mathbb{R}^3 $, the rays follow a linear relationship
\begin{equation}
  \begin{bmatrix}
    u \\
      v
  \end{bmatrix}
  = \begin{pmatrix}
    \frac{D}{P_{z}}
    \end{pmatrix}
    \begin{bmatrix}
      P_{x} - s \\
      P_{y} - t
    \end{bmatrix},
  \label{eq:ptPlane4D}
\end{equation}
where each of these equations describes a hyperplane in 4D.
%In the LF literature, geometrical definitions of hyperplanes are often not clearly stated.
In this paper, a hyperplane is defined as a vector subspace that has 1 dimension less than the space it is contained within~\cite{hyperplane}. Thus a hyperplane in 4D is a 3-dimensional manifold, and can be described by a single equation
% \dorian{where are these 3 dimensions in this equation? Normal is 4D, so $stuv$ are constrained to 3 dimensions?}
\begin{equation}
  n_1 s + n_2 t + n_3 u + n_4 v + n_5 = 0,
  \label{eq:hyperplaneAlgebraic}
\end{equation}
where $\vec{n} = [n_1, n_2, n_3, n_4]^T$ is the normal of the hyperplane. Similarly, a plane is defined as a 2-dimensional manifold; in other words, it can be described by two linearly independent vectors. Therefore, a plane in 4D can be defined by the intersection of two hyperplanes
\begin{comment}
\begin{align}
& n_1 s + n_2 t + n_3 u + n_4 v + n_5 = 0\\
& m_1 s + m_2 t + m_3 u + m_4 v + m_5 = 0,
\label{eq:planeAlgebraic}
\end{align}
where $m$ is the normal of a second hyperplane in 4D.
\eq{eq:planeAlgebraic} can then be re-organized into matrix form,
\begin{align}
\begin{bmatrix}
n_1 & n_2 & n_3 & n_4 \\
m_1 & m_2 & m_3 & m_4
\end{bmatrix}
\begin{bmatrix}
s\\t\\u\\v
\end{bmatrix}
=
\begin{bmatrix}
-n_5 \\ -m_5
\end{bmatrix}.
\label{eq:plane4D}
\end{align}
\end{comment}
and (\ref{eq:ptPlane4D}) can be re-written in the form,
\begin{align}
\underbrace{
\begin{bmatrix}
\frac{D}{P_z} & 0 & 1 & 0 \\
0 & \frac{D}{P_z} & 0 & 1
\end{bmatrix}
}_\text{$\vec{m}$}
\begin{bmatrix}
s\\t\\u\\v
\end{bmatrix}
=
\begin{bmatrix}
\frac{D P_x}{P_z} \\ \frac{D P_y}{P_z}
\end{bmatrix},
\label{eq:plane4D_2PP}
\end{align}
where $\vec{m}$ contains the two orthogonal normals to the plane.
Therefore, a Lambertian point in 3D manifests itself as a plane in 4D, which is characterized by two linearly-independent normal vectors that each define a hyperplane in 4D. In the literature, this relationship is sometimes referred to as the point-plane correspondence.

% \subsection{Light Field Slope}

Light field slope $w$ relates the rate of change of image plane coordinates, with respect to viewpoint position, for all rays emanating from a point in the scene. In the literature, slope is sometimes referred to as ``orientation''~\cite{wanner2013reflectiveTransparentFromEPI}, and other works compute slope as an angle~\cite{tosic2014lightFieldScaleSpace}. % Fig.~\ref{fig:lightfieldSlopeGeometry} shows the 2D view of the geometry of a single point in the LF as several rays intersecting the two planes at different values of $s$ and $u$. While holding $t$ and $v$ constant, as the viewpoint $s$ changes, the image plane coordinate $u$ varies linearly according to Eq.~\ref{eq:ptPlane4D}. Fig.~\ref{fig:lightfieldSlopeGeometry} shows how $u$ varies with $s$, noting that $v$ similarly varies with $t$.
The slope comes directly from (\ref{eq:ptPlane4D}) as
\begin{equation}
w = -\frac{D}{P_z},
\end{equation}
and is clearly related to depth. Importantly, slope is uniform; it is identical for the $s,u$ and $t,v$ planes for a Lambertian point.

\begin{comment}
\begin{figure}[htbp]
\centering
\subfloat[][]{\includegraphics[width=0.25\textwidth]{Figures/LightfieldGeometry-Dansereau.png}\label{fig:lightfieldGeometry}}\hfil
\subfloat[][]{\includegraphics[width=0.2\textwidth]{Figures/LightfieldGeometrySlope-Dansereau}\label{fig:lightfieldSlope}}
\caption{(a) Light field geometry for a point in space for a single view (black), and other views (grey), whereby $u$ varies linearly with $s$ for all rays originating from $^c{P}$. (b) The corresponding line in the $s,u$ plane, having the slope $w$~\cite{dansereau2014Thesis}.}
\label{fig:lightfieldSlopeGeometry}
\end{figure}
\end{comment}

%%%%%%%%%%%%%%%%%%%%%%%%%%%%%%%%%%%%%%%%%%%%%%%%%%%%%%%%%%%%%%%%%%%%%%%%%%%%%%%%
\section{DISTINGUISHING REFRACTIVE FEATURES}

% \subsection{Epipolar Plane Images in the Light Field}

\begin{figure}
\centering
\subfloat[][]{\includegraphics[width=0.38\textwidth]{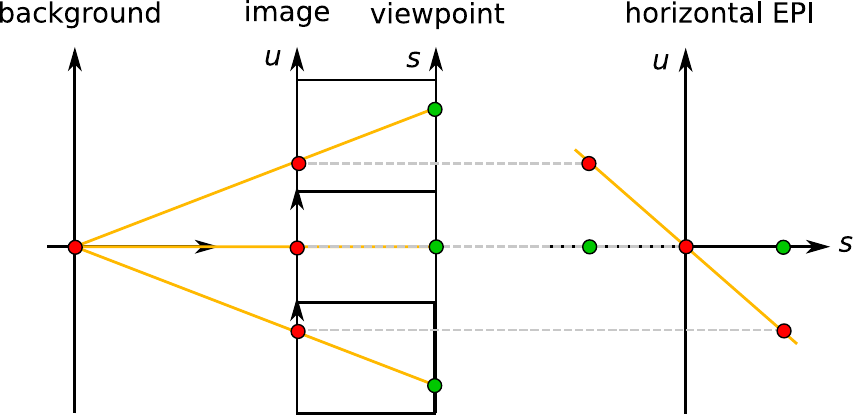}
\label{fig:XuLinear}}\\
\subfloat[][]{\includegraphics[width=0.38\textwidth]{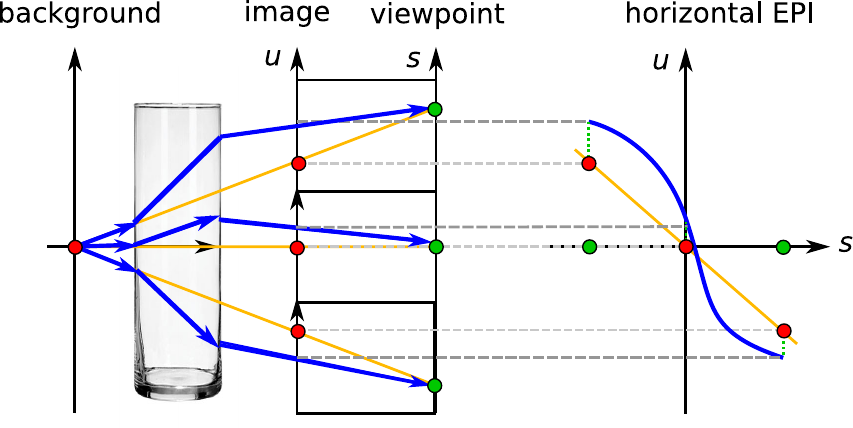}
\label{fig:XuTransparent}}\\
\caption{(a) Projection of the linear behaviour of a Lambertian feature, and (b) the non-linear behaviour of a refracted feature with respect to linear motion along the viewpoints of a light field.}
\label{fig:FeatureBehaviour}
\vspace{-1.5em}
\end{figure}

Epipolar planar images (EPIs) graphically illustrate the apparent motion of a feature across multiple views~\cite{bolles1987epipolar}.
If the entire light field $L$ is given as $L(s,t,u,v)$, EPIs represent a 2D slice of the 4D LF. A horizontal EPI is given as $L(s,t^*,u,v^*)$, and a vertical EPI is denoted as $L(s^*,t,u^*,v)$, where $^*$ indicates a variable is fixed while others are allowed to vary. The central view of the LF is given as $L(s_0,t_0,u,v)$, and is equivalent to what a monocular camera would provide from the same pose.
As shown in Fig.~\ref{fig:FeatureBehaviour}, features from a Lambertian scene point are linearly distributed with respect to viewpoint, unlike features from highly-distorting refractive objects. We compare this difference in apparent motion between Lambertian and non-Lambertian features to distinguish refracted features.

% Show example EPI, probably in conjunction with sample image and LF notation.

\begin{figure}[tbp]
\centering
\subfloat[][]{\includegraphics[height=0.6\columnwidth]{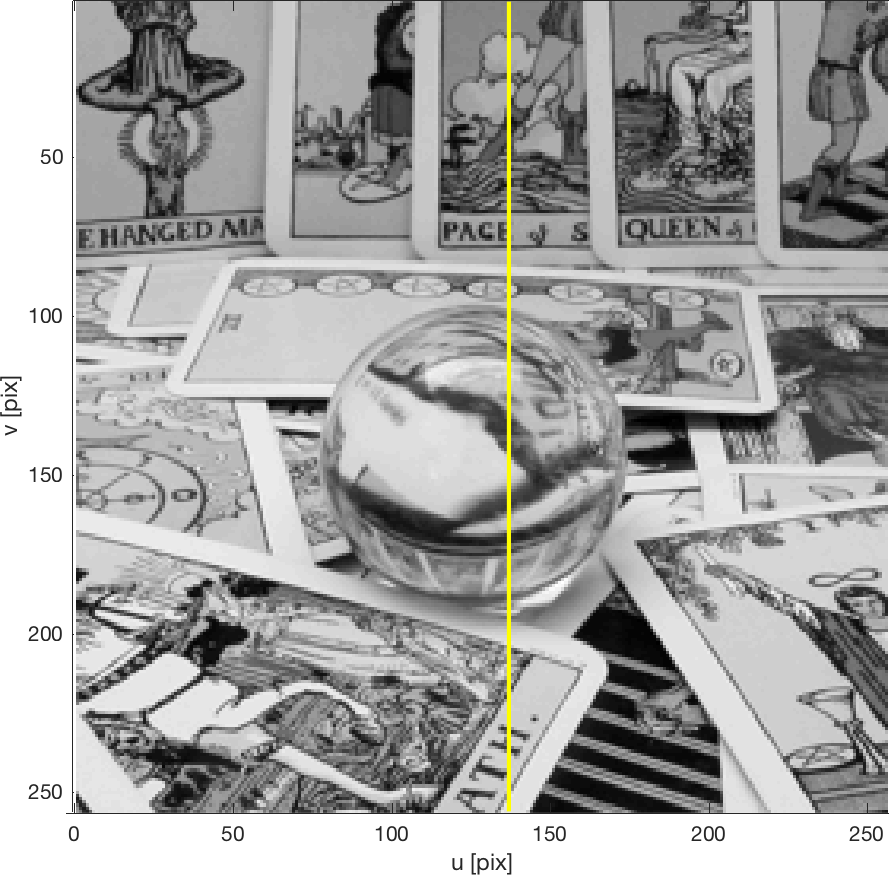}\label{fig:lightfieldSlope}}\hfil
\subfloat[][]{\includegraphics[height=0.6\columnwidth]{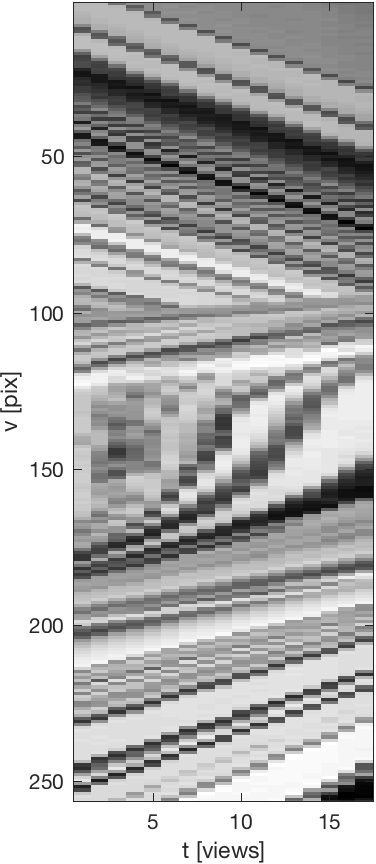}\label{fig:exVertEpi}}
\caption{
(a) In the crystal ball LF~\cite{stanfordLightfieldArchiveNew}, a vertical EPI is sampled from a column of pixels (yellow), where (b) nonlinear refracted feature motion caused by the crystal ball are apparent in the middle of this vertical EPI. Straight lines from $v < 100$, and $v > 200$ relate to Lambertian features.
%(a) The central view of the LF with the yellow line indicating where the EPI was taken. Original image down-sampled for computation~\cite{stanfordLightfieldArchiveNew}. (b) Vertical EPI showing straight lines for the Lambertian features for $v < 100$, and $v > 200$ pix, and non-linear curves for the refractive sphere in between.
}
\label{fig:exEpi2}
\vspace{-1.0em}
\end{figure}

Fig.~\ref{fig:exEpi2} shows the central view and an example EPI of a crystal ball LF (large baseline) from the New Stanford Light Field Archive~\cite{stanfordLightfieldArchiveNew}. A Lambertian point forms a straight line in the EPI, which represents one of the point's hyperplanes in 4D, as illustrated by the Lambertian scene content in Fig.~\ref{fig:exVertEpi}, i.e. to the top ($v < 100$), and bottom ($v > 200$) of the crystal ball. The relation between slope and depth is also apparent in this EPI.
% EPIs can also be taken about different viewpoint-axes in $s$ and $t$. Each of these represents a different hyperplane, but
% A minimum representation of two linearly-independent hyperplanes are required to describe a 3D point $\vec{P}$ on a Lambertian surface as it appears in the 4D LF.

For a refracted feature, such as those seen in Fig.~\ref{fig:exVertEpi} for $100 < v < 200$ pix, detection in the LF simplifies to finding features that violate (\ref{eq:ptPlane4D}) via identifying non-linear feature curves in the EPIs and/or inconsistent slopes between two independent EPI lines, such as the vertical and horizontal EPIs.
We note that occlusions and specular reflections also violate (\ref{eq:ptPlane4D}). Occlusions appear as straight lines, but have intersections in the EPI. Edges of the refractive objects, and objects with low distortion also appear Lambertian. Specular reflections appear as a superposition of lines in the EPI. We will address these issues in future work. For now, we discuss how we extracted these 4D feature curves, and then describe how we identify refracted features.

% For a refractive feature, the concept of a single slope is not valid.

%%%%%%%%%%%%%%%%%%%%%%%%%%%%%%%%%%%%%%%%%%%%%%%%%%%%%%%%%%%%%%%%%%%%%%%%%%%%%%%%

\subsection{Extracting Feature Curves}

For a given feature from the central view at coordinates $(u_0, v_0)$, we must determine the feature correspondences $(u^\prime,v^\prime)$ from the other views, which is equivalent to finding the feature's apparent motion, or curves in the LF. In this paper, we start by detecting SIFT features~\cite{lowe2004distinctive} in the central view, although the proposed method is agnostic to feature type.
% , provided they are sufficiently well-defined for the 2D autocorrelation function. % not exactly - edge features don't bode well - any features that play well with auto-correlation
% why we don't just find edges
% nEdge detection methods were considered; however, good features to track, such as those provided by SIFT do not necessarily lie precisely on edges within the EPI.
% How we extract characteristic feature curves using textural autocorrelation. State/describe how certain thresholds are tailored to the application space, and try to enforce line consistency from current max of the NCC-EPI.

\begin{figure}[tbp]
\centering
\subfloat[][]{\includegraphics[width=0.45\columnwidth,height=0.15\columnwidth]{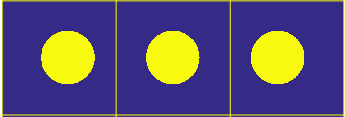}\label{fig:views}}\hfil
\subfloat[][]{\includegraphics[width=0.45\columnwidth,height=0.15\columnwidth]{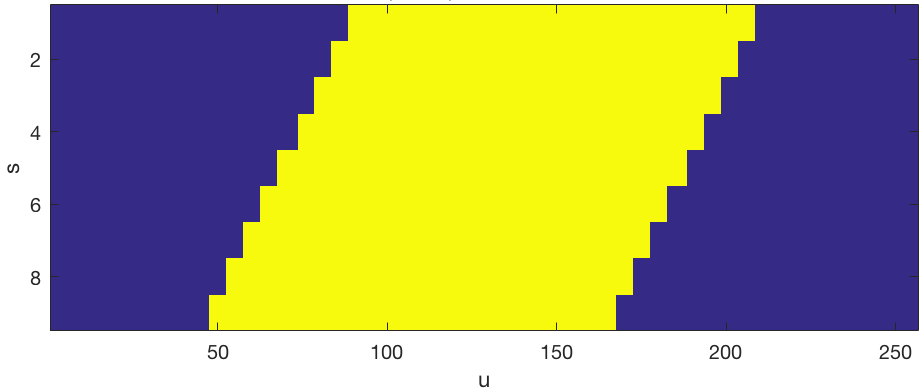}\label{fig:epiRegular}}\\
\subfloat[][]{\includegraphics[width=0.45\columnwidth,height=0.15\columnwidth]{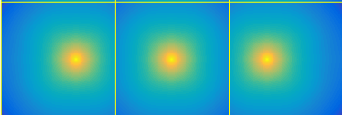}\label{fig:viewCorr}}\hfil
\subfloat[][]{\includegraphics[width=0.45\columnwidth,height=0.15\columnwidth]{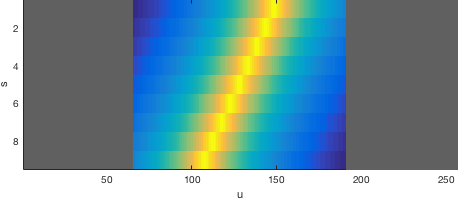}\label{fig:epiCorr}}
\caption{(a) Simulated horizontal views of a yellow circle. (b) Corresponding EPI taken along the middle of the views. (c) The cross-correlation response for the corresponding views. (d) The resultant correlation EPI, created by stacking the cross-correlation response from adjacent views. The peaks of this correlation EPI correspond to the desired feature curve.}
\label{fig:correlation}
\vspace{-1.0em}
\end{figure}

Next, we select a template surrounding the feature which is $k$-times the feature's scale. We determined $k=5$ to yield the most consistent results. 2D Gaussian-weighted normalized cross-correlation (WNCC) is used across views to yield images, such as Fig.~\ref{fig:viewCorr}. To reduce computation, we only apply WNCC along a subset of the central row and column relative to the central view of the LF.

% a feature template is defined as the square neighbourhood of size $k$-times the feature's scale, where $k>0$ is a tunable parameter. Currently, 2D Gaussian-weighted normalized cross-correlation (WNCC) is used across only the views associated with the given EPI. Thus all views of $s$ for $t_0$, and all $t$ for $s_0$ are currently used.
%This forms a correlation space that represents a minimal sampling along the horizontal and vertical epipolar lines.
% The correlation could be applied to every view for more sampling at the cost of more complex computation. To reduce computation, WNCC is only performed on the image rows covered by the template, centred about the given horizontal EPI (or columns for the vertical EPI).
%  Note: wncc is good when scale, orientation is known. I don't use SIFT's orientation info! Also, need to reference Andrew Diamond's wncc implementation?}

For Lambertian features, %peaks in the correlation space for each view correspond to the feature's image coordinates in that view. Thus
we plot the feature's correlation response with respect to the views to yield a correlation EPI. Illustrated in Fig.~\ref{fig:epiCorr}, the peaks of the correlation EPI correspond to the feature curve from original EPI.
% Extracted curves from a single EPI provide a single set of 4D points (for example, we find $s,u$ from a horizontal EPI with constant $t^*,v^*$), that correspond to a hyperplane in 4D for a Lambertian point.

For refracted features, we hypothesize that the correlation response will be sufficiently strong that peak values of the correlation EPI will still correspond to the desired feature curve.
% \dorian{Extracted yellow line is not perfectly smooth. Consider smoothing correlation EPI to reduce the anti-aliasing effects of limited angular resolution.}
As such, we threshold the correlation EPI into a binary mask to reduce the area considered for the feature curve. We then take the peak values of the connected component that contains our feature. We apply several line consistency thresholds to remove incorrectly-detected feature curves due to edge boundaries or peak values corresponding to the wrong features. Thresholds were tailored for the specific application. This textural cross-correlation method allows us to focus on the image structure, as opposed to the image intensities, can be applied to any LF camera, and directly exploits the geometry of the LF.

% \dorian{Can potentially improve consistency of match by checking forwards/backwards in EPIs or with actual views}

\subsection{Fitting 4D Planarity to Feature Curves}

Similar to \cite{xu2015transparentObjectSegmentation}, we consider the feature ray $\phi(0,0,u_0,v_0)$. The corresponding feature coordinates in other views are $\phi^\prime(s,t,u^\prime,v^\prime)$. The LFD is then defined as the set of relative differences between $\phi$ and $\phi^\prime$:
\begin{equation}
LFD(u,v) = \{(s,t,\Delta u, \Delta v) | (s,t) ~= (0,0)\},
\label{eq:lfd}
\end{equation}
where $\Delta u = u^\prime - u_0$, and $\Delta v = v^\prime - v_0$ are feature disparities.

As illustrated in Fig.~\ref{fig:XuLinear}, these disparities are linear with respect to linear camera translation. The disparities from refracted features deviate from this linear relation. Fitting them to (\ref{eq:ptPlane4D}) yields the plane of best fit in 4D, and the error of this fit provides a measure of whether or not our feature is Lambertian.

% While Maeno and Xu used all $s$ and $t$, we only use $s$ and $t$ relative to the central view. These values are the feature curves from our EPIs (previous section).
This plane in 4D can be estimated from the feature correspondences given by the feature curves $f_h(s,t^*,\Delta u,v^* - v_0)$, and $f_v(s^*,t,u^* - u_0,\Delta v)$ that we extract from the horizontal and vertical EPIs, respectively.
% \dorian{An interesting question is minima cross-representation (just horz/vert EPI) vs entire LF?}

As discussed in Section III, our plane in 4D has two orthogonal normals, $\vec{n}_h$ and $\vec{n}_v$.
The 4D plane containing $\phi$ can be given as
\begin{align}
\begin{bmatrix}
n_{h,1} & n_{h,2} & n_{h,3} & n_{h,4} \\
n_{v,1} & n_{v,2} & n_{v,3} & n_{v,4}
\end{bmatrix}
\begin{bmatrix}
s\\t\\ \Delta u\\ \Delta v
\end{bmatrix}
=
\begin{bmatrix}
0 \\ 0
\end{bmatrix}.
\end{align}
Note that the constants on the right-hand side of (\ref{eq:plane4D_2PP}) cancel out because we consider the differences relative to $u_0$ and $v_0$.
The positions for $s,t$ can be obtained by calibration~\cite{dansereau2013decoding}, although the non-linearity behaviour still holds when working with units of ``views'' for $s,t$.

We can estimate $\vec{n}_h$ and $\vec{n}_v$ by fitting the $N$ points from $f_h$ and $M$ points from $f_v$ into the system,
\begin{align}
\underbrace{
\begin{bmatrix}
(s, & t^*, & \Delta u, & v^*-v_0)_1 \\
\vdots & \vdots & \vdots & \vdots \\
(s, & t^*, & \Delta u, & v^*-v_0)_N \\
(s^*, & t, & u^*-u_0, & \Delta v)_1 \\
\vdots & \vdots & \vdots & \vdots \\
(s^*, & t, & u^*-u_0, & \Delta v)_M \\
\end{bmatrix}
}_\text{$\vec{A}$}
\underbrace{
\begin{bmatrix}
n_1\\
n_2\\
n_3\\
n_4
\end{bmatrix}
}_\text{$\vec{n}$}
=
\vec{0}.
\label{eq:svd}
\end{align}

We then use SVD on $\vec{A}$ to compute the singular vectors, and corresponding singular values. The 2 smallest singular values, $e_1$ and $e_2$, correspond to 2 normals $\vec{n}_1$ and $\vec{n}_2$ that best satisfy (\ref{eq:svd}) in the least-squares sense. The magnitude of the singular values provides an error measure of the planar fit. Smaller errors imply stronger linearity, while larger errors imply that the feature deviates from the 4D plane.

% We can identify which singular values correspond to the horizontal and vertical EPIs by considering the relative values of $\vec{n}_1$ and $\vec{n}_2$.
% For the horizontal EPI, $\vec{n}_h$ closely resembles the first normal in (\ref{eq:plane4D_2PP}), while $\vec{n}_v$ closely resembles the second normal in (\ref{eq:plane4D_2PP}).

The norm of $e_1$ and $e_2$ may be taken as a single measure of planarity; however, doing so masks the case where a refractive object has unequal errors between the two EPIs, such as a 1D refractive object (glass cylinder) that is highly refractive along one direction, but almost Lambertian along the other. Therefore, we reject those features that have large errors in either horizontal or vertical hyperplanes, according to an empirical threshold. % This is what makes us more sensitive to different types of refractive objects, which was not accounted for in Xu's work.
This planar consistency, along with the slope consistency measure discussed in the following section, make the proposed method more sensitive to refracted texture than prior work that considers only hyperplanar consistency~\cite{xu2015transparentObjectSegmentation}.

% Xu's approach samples from all the views in the LF, but only considers the smallest singular value as a measure of how Lambertian the target feature appears.
While Xu also applies occlusion handling and a graph segmentation framework to complete their transparent object image segmentation algorithm, simply taking the smallest singular value allows us to directly compare the underlying principles for our feature distinguisher.

\subsection{Measuring Slope Consistency}

Slope consistency is a measure of how similar the slopes are between the two hyperplanes for a given feature. As seen in (\ref{eq:ptPlane4D}), these slopes must be equal for Lambertian points. We can compute the slopes for each hyperplane given their normals. For the horizontal hyperplane, we solve for in-plane vector $\vec{q} = [q_s, q_u]^T$, by taking the inner product of the two vectors in
\begin{align}
\begin{bmatrix}
n_{h,1} & n_{h,3} \\
n_{v,1} & n_{v,3}
\end{bmatrix}
\begin{bmatrix}
q_s \\
q_u
\end{bmatrix}
 = \vec{0},
\end{align}
where $\vec{q}$ is constrained to the $s,u$ plane, because we choose the first and third elements of $\vec{n}_h$ and $\vec{n}_v$. This system is solved using SVD, and the minimum singular vector yields $\vec{q}$. The slope for the horizontal hyperplane, $w_{su}$ is then
\begin{equation}
w_{su} = \frac{q_s}{q_u}.
\end{equation}
The slope for the vertical hyperplane $w_{tv}$ is similarly computed from the second and fourth elements of the normal vectors.
Alternatively, one can also find the line of best fit for the $s,u$-values in $f_h$, and the $t,v$-values in $f_v$.
% In practice, one can also take the negative reciprocal of the hyperplane's normal for the given $s,u$ or $t,v$ planes.
Slope inconsistency $c$ is calculated as the square of differences between slopes.
%\begin{equation}
%c = (w_{su} - w_{tv})^2.
%\end{equation}
Finally, a threshold for slope inconsistency is applied, which is tuned for the application. Features with very inconsistent slopes and large planar errors are identified as belonging to a highly-distorting refractive object, which we refer to as a refracted feature.

% Thanks to our selection of SIFT features (that lie on blobs), occlusions (that lie on edges) are not often detected.

%%%%%%%%%%%%%%%%%%%%%%%%%%%%%%%%%%%%%%%%%%%%%%%%%%%%%%%%%%%%%%%%%%%%%%%%%%%%%%%%

\section{EXPERIMENTAL RESULTS}

% outline of what results we will be discussing:
First, we present our experimental set-up. Second, we present results of our feature extraction method for Lambertian and refracted features. Third, we apply our methods to LFs captured with different baselines.
%, in order to evaluate our performance in distinguishing refractive features with respect to baseline.
Fourth, we apply our methods to LFs captured with a lenslet-based LF camera to compare different refractive object types. Finally, we use our method to reject refracted features for monocular SfM in the presence of refractive objects, and investigate the impact of our approach.

\subsection{Experimental Set-up}

We used the Stanford New Light Field Database~\cite{stanfordLightfieldArchiveNew}, which provided LFs captured from a Lego gantry with a $17\times17$ grid of rectified $1024\times1024$-pixel images that were down-sampled to $256\times256$ pixels to reduce computation. We focused on two LFs that both captured the same scene of a crystal ball surrounded by textured tarot cards. The first was captured with a large baseline (16.1 mm/view over 275 mm),
% because of the significant distortions in the LF caused by the crystal ball.
while the second was captured with a smaller baseline (3.7 mm/view over 64 mm). This allowed us to compare the effect of LF camera baseline for refracted features.

% Describe experiments, key results in plots, discussion of plots.
Even smaller baselines were considered using a lenslet-based LF camera. Also known as a plenoptic camera, these LF cameras are of interest to robotics due to their simultaneous view capture, and typically lower size and mass, compared to LF camera arrays and gantries. In this section, the Lytro Illum was used to capture LFs with $15\times15$ views, each $433\times625$ pixels. Dansereau's Light Field Toolbox was used to decode the LFs from raw LF imagery to the 2PP~\cite{dansereau2013decoding}. 
% can remove the entire next two paragraphs if short on space!
To compensate for the extreme lens distortion of the Illum, we removed the outer views, reducing our LF to $13\times13$ views. The LF camera was fixed at 100 mm zoom. %, a trade-off between plausible manipulation distance, extreme lens distortions observed at the extreme zoom settings, and vignetting. 
For these optics, the Illum was roughly equivalent to a 1.1 mm/view over 16.6 mm LF camera array. 
% Since the refractive object was placed close to the background, the Illum was focused on the Lambertian background when LFs were captured from different distances.
% These non-constant focus settings were not a significant issue for our experiments, because slope was not being used to recover depth between multiple LFs. All LFs were captured in ambient indoor lighting conditions without the need for specialized lighting equipment.

It is important to remember that our results depend on a number of factors. First, the shape and size of the object dictates how the light is refracted. Higher curvature and thickness yield more warping. Second, viewing distance, and background distance to the object directly affect how much distortion can be observed. The closer either is to the object, the more refraction can be observed. Similarly, a larger camera baseline captures more refraction. When possible, these factors were held constant throughout different experiments.

\subsection{Feature Curve Extraction}

\begin{figure}[t]
\centering
% \subfloat[][]{\includegraphics[height=0.35\columnwidth]{Figures/LambertianTemplate.png}\label{fig:lambTemplate}}\\
\subfloat[]{\includegraphics[width=0.5\columnwidth]{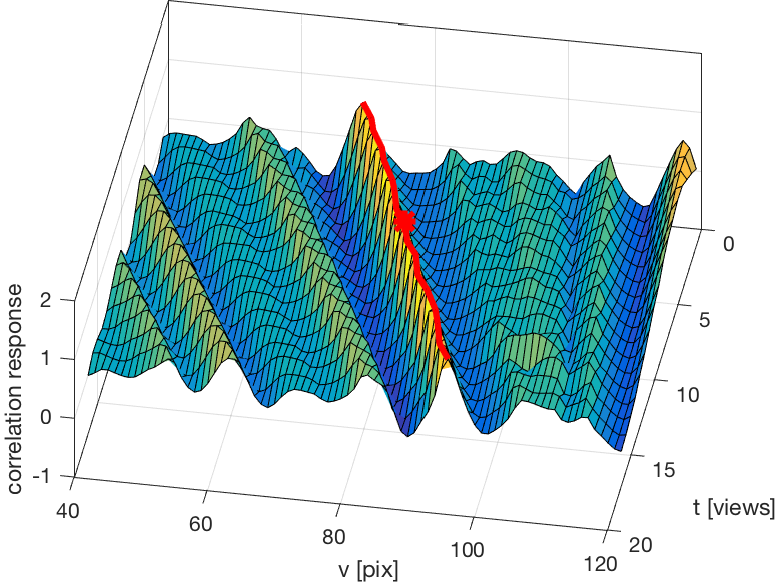}\label{fig:lambCorrelationEpi}}\hfil
\subfloat[]{\includegraphics[width=0.5\columnwidth]{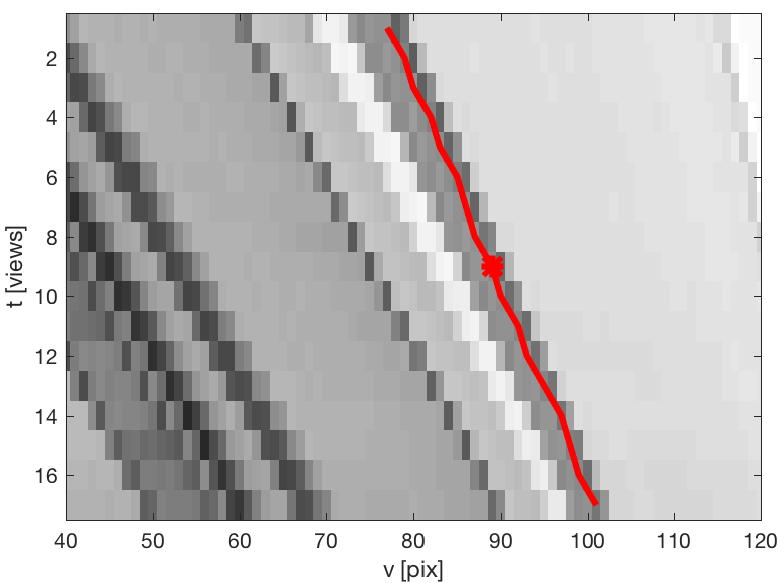}\label{fig:lambFeatureCurve}}\\
% \subfloat[][]{\includegraphics[height=0.35\columnwidth]{Figures/crystalLarge_Templace.png}\label{fig:nccTemplate}}\\
% \subfloat[][]{\includegraphics[width=0.45\textwidth]{Figures/crystalLarge_NCCEpiSurface.png}\label{fig:nccEpiSurface}}\\
\subfloat[]{\includegraphics[width=0.5\columnwidth]{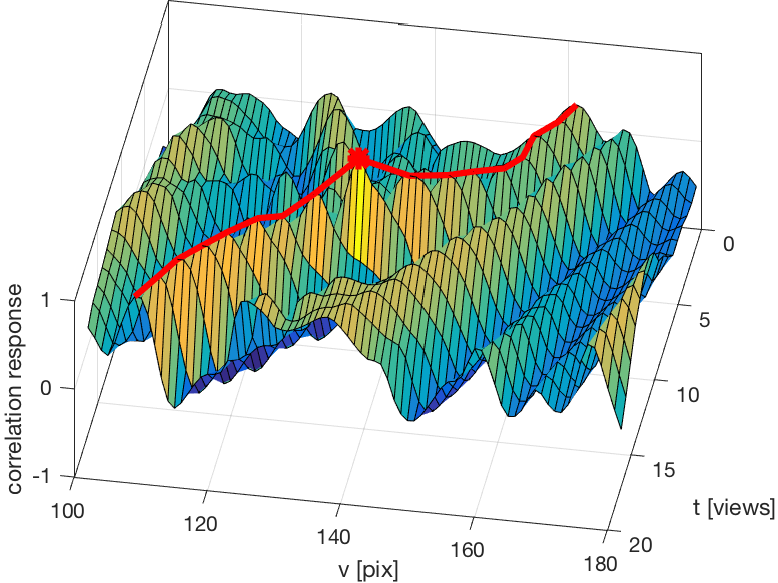}\label{fig:nccEpiFlat}}\hfil
\subfloat[]{\includegraphics[width=0.5\columnwidth]{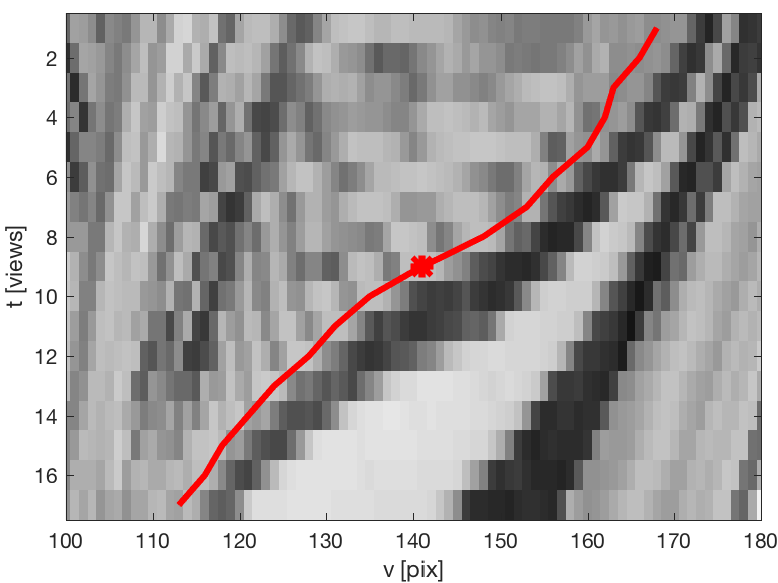}\label{fig:nccLine}}
\caption{% (a) Sample Lambertian SIFT feature with template indicated in red, which is $k$ times the feature's scale.
(a) The straight Lambertian feature curve (red) from the correlation EPI corresponds to (b) the feature curve in the original EPI.
(c) The refracted feature curve (red) in the correlation EPI can still be extracted, despite more complex ``terrain'', and still matches (d) the curve just above the refracted black curve in the original EPI.
%(a) Lambertian SIFT feature's correlation EPI, shown as a surface and the extracted feature curve shown as a straight red line.
%(b) The extracted curve from the correlation EPI corresponds to the feature's curve in the normal EPI with with feature's central view location as the red dot, and extracted curve in red.
% (a) Refractive feature template.
%(c) Refractive SIFT feature's correlation EPI, showing significantly more ridges and varied terrain, but the feature curve can still be extracted from the local peak values, in red.
%(d) Extracted feature curve in red, super-imposed on the normal EPI, shown to correspond with the feature just above the refracted black curve.
}
\label{fig:extractFeatureCurves}
\vspace{-1.0em}
\end{figure}

We first considered our textural cross-correlation approach, which worked well for Lambertian features. For such a feature % shown in Fig.~\ref{fig:lambTemplate},
the correlation EPI is shown as a surface in Fig.~\ref{fig:lambCorrelationEpi} with the feature from the central view shown as the red dot. The corresponding peaks from the other views are shown as the red line. This line corresponds to the feature curve in the image space, as shown in Fig.~\ref{fig:lambFeatureCurve}. Line consistency thresholds were implemented to ensure consistent feature curves.
Although not implemented, we can limit the cross-correlation space, since we know the features move with some maximum slope, to further reduce computation.

For refracted features, our approach captured the feature curves. For a typical refracted SIFT feature, % shown in Fig.~\ref{fig:nccTemplate}
the correlation EPI is shown in Fig.~\ref{fig:nccEpiFlat}. We observed that while the correlation function often had a much weaker response compared to the Lambertian case, local maxima were still observed near the original feature's spatial location in the central view. Thus, taking the local maxima of the correlation EPI still yielded the desired feature curve, as shown in Fig.~\ref{fig:nccLine}. 
% \dorian{consider adding problem cases to showcase difficulty/limits of thresholds}

% ommitted for space
% The refractive feature curves were more difficult to obtain. The significant distortions from the refractive object significantly varied the correlation response, such that applying a threshold and taking the peak values occasionally yielded inconsistent feature curves. Line consistency thresholds were implemented to remove inconsistent portions of the feature curves; however, shorter curves were more easily fit to straight lines, which then caused the curve to be distinguished as a straight Lambertian line.

Our method relies on thresholds that we have set empirically, which results in limited flexibility to changing environmental conditions. As future work, we are considering implementing ridge-following methods to improve the feature curve extraction.
%An alternate method for feature curve extraction would be to follow the ridge of the correlation EPI, starting from its peak. However, this approach requires Hessians from the heavily-aliased EPI data.  Smoothing and up-sampling may improve the anti-aliasing caused by the relatively large baseline between views.
Nonetheless, our textural cross-correlation method enabled us to extract refracted feature curves without focusing on image intensities in a way that exploited the geometry of the LF.

\subsection{Refracted Feature Comparison for LF Camera Baseline}

% \dorian{true positives, false negatives, false positives, true negatives, what is ``good'' for this binary classifier, under what conditions, consider subsubsection}
For implementation, the thresholds for planarity and slope consistency were manually applied in order to provide optimal results for both Xu's and our approach, independently.
For metrics, the number of refracted feature detections and false positives were counted, and compared. The true positive rate (TPR) was given as the number of refracted features detected in the central view within the image space occupied by the refractive object, over the number of actual refracted features, obtained via a mask identified by the author for the refractive object in the central view. The false positive rate (FPR) was calculated as the number of features falsely identified as refracted over the number of features not refracted.

\begin{figure}[t]
\centering
\scriptsize
\setlength\tabcolsep{1.5pt} % default 6pt
\begin{tabular}{ccc}
%\hline
	& Xu's Method & Our Method \\
%\hline
\raisebox{0.16\columnwidth}{\rotatebox[origin=c]{90}{Large baseline, crystal ball}}
	& \includegraphics[width=0.47\columnwidth,height=0.34\columnwidth]{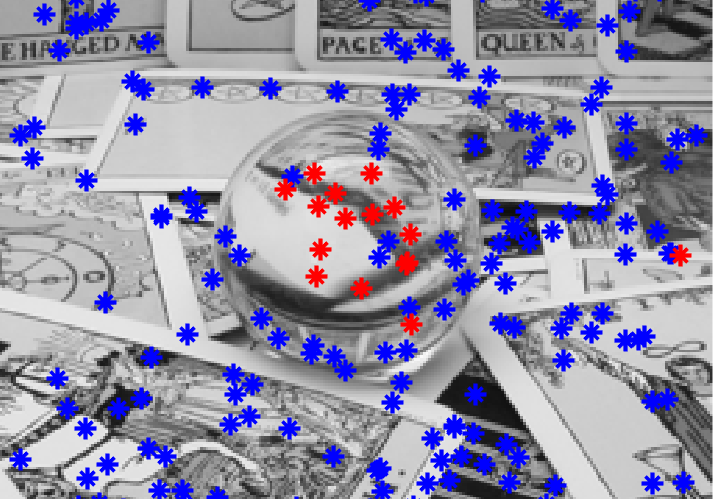}
	%\subfloat[]{\includegraphics[width=0.47\columnwidth]{Figures/LFFeatures_XuHyperplanarityThreshcrystalLarge.png}\label{fig:crystalLargeXu}}
	%\vspace{-0.5em}
	& \includegraphics[width=0.47\columnwidth,height=0.34\columnwidth]{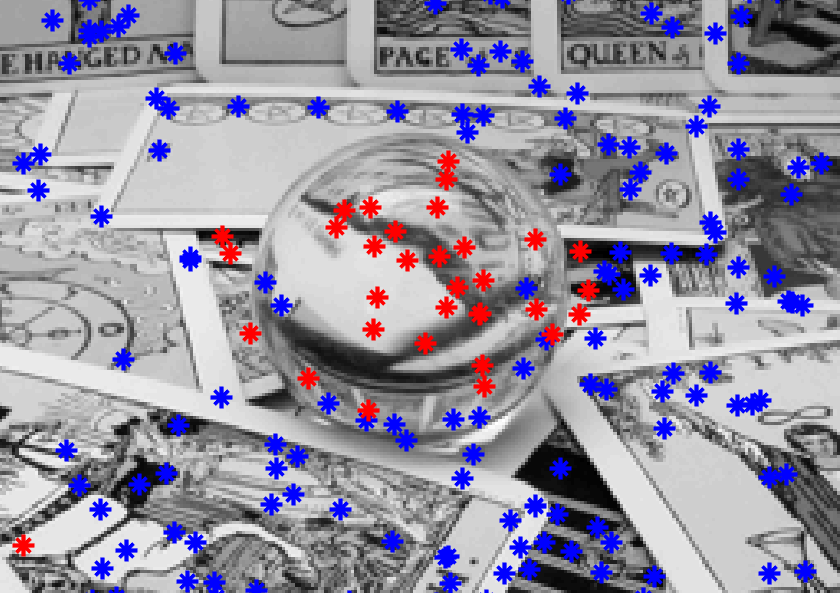}
	%\vspace{-0.5em}
	\\
\raisebox{0.16\columnwidth}{\rotatebox[origin=c]{90}{Small baseline, cylinder}}
	& \includegraphics[width=0.47\columnwidth]{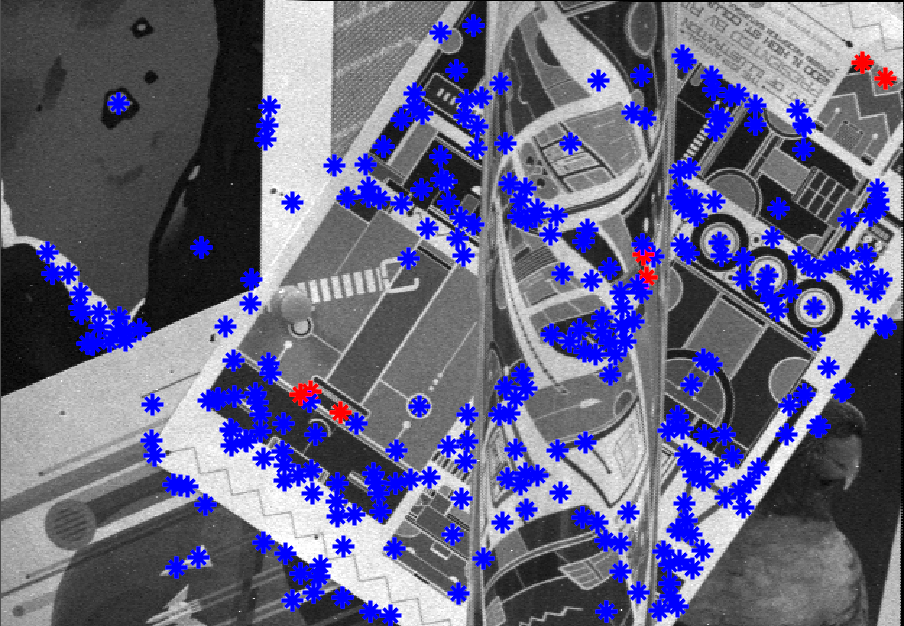}
	%\vspace{-0.5em}
	& \includegraphics[width=0.47\columnwidth]{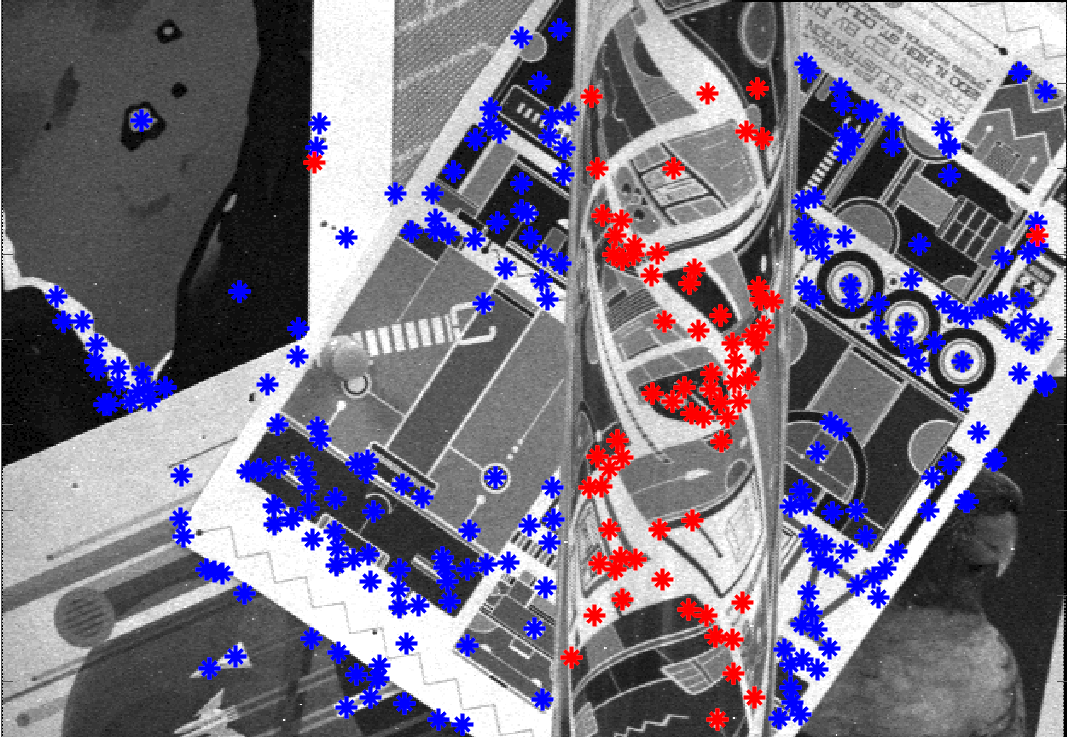}
	%\vspace{-0.5em}
	\\
\raisebox{0.16\columnwidth}{\rotatebox[origin=c]{90}{Small baseline, sphere}}
	& \includegraphics[width=0.47\columnwidth]{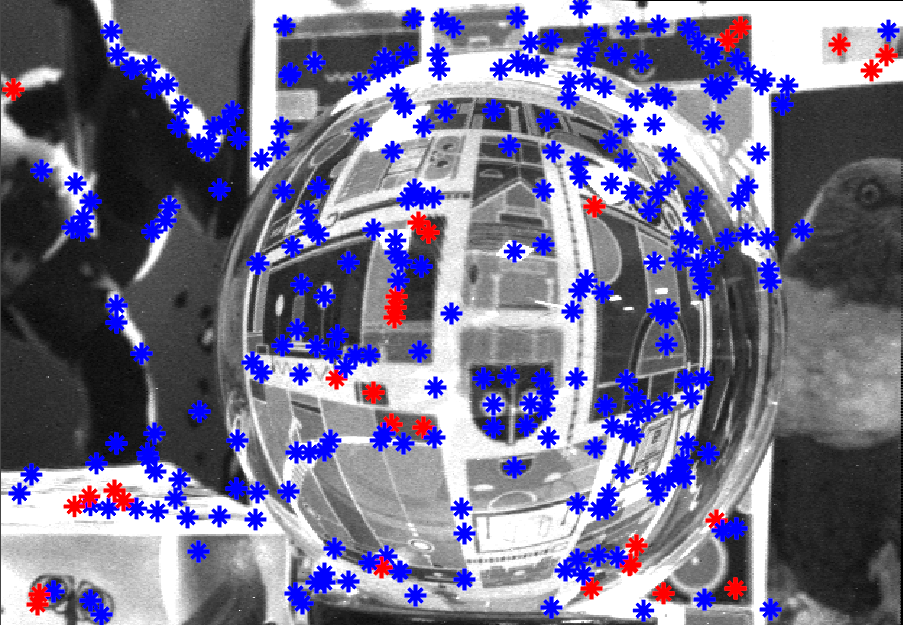}%\label{fig:XusphereRefractive}}
	%\vspace{-0.5em}
	& \includegraphics[width=0.47\columnwidth]{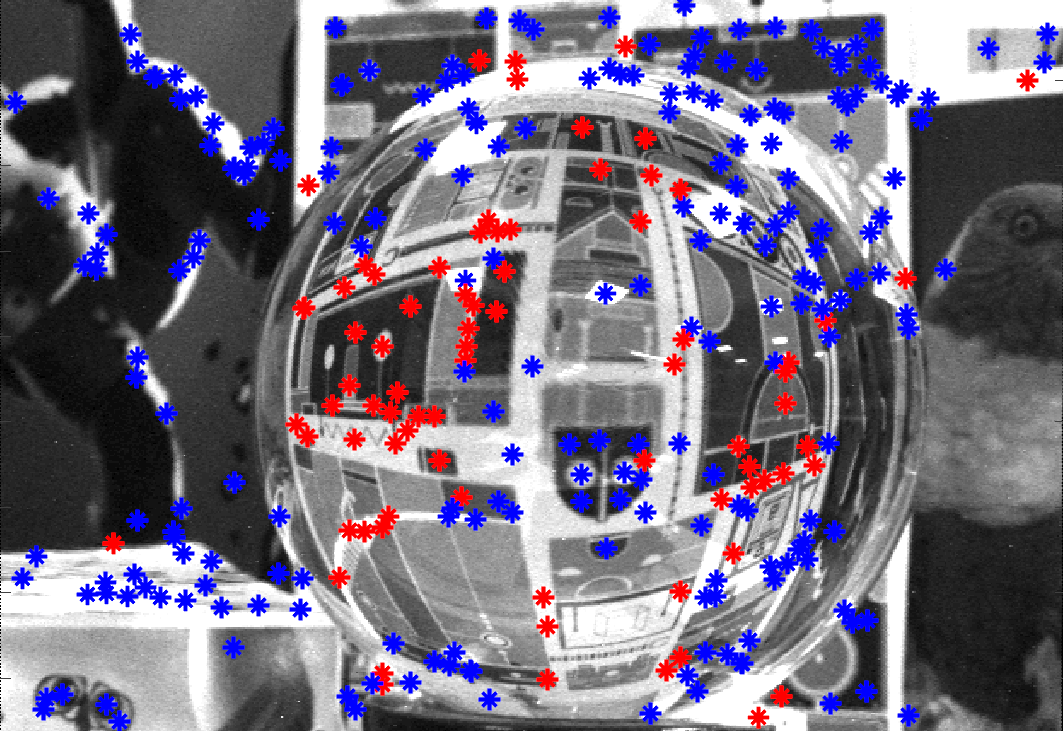}%\label{fig:DdsphereRefractive}}
	%\vspace{-0.5em}
	\\
%\hline
\end{tabular}
\caption{Comparison of Xu's method (left), and our method (right), detecting Lambertian (blue), and refracted (red) SIFT features.
In the top row, the crystal ball captured with a large baseline LF is shown (cropped)~\cite{stanfordLightfieldArchiveNew}, where both methods detect refracted features; however, our method outperforms Xu's.
In the second and third rows, a cylinder and sphere captured with a small-baseline lenslet-based LF camera. Our method successfully detects more refracted features with fewer false positives, while Xu's method does not reliably detect refracted features for small baselines.
}
\vspace{-1.0em}
\label{fig:comparison}
\end{figure}

\begin{figure}[t]
\centering
\subfloat[][]{\includegraphics[width=0.49\columnwidth,height=0.37\columnwidth]{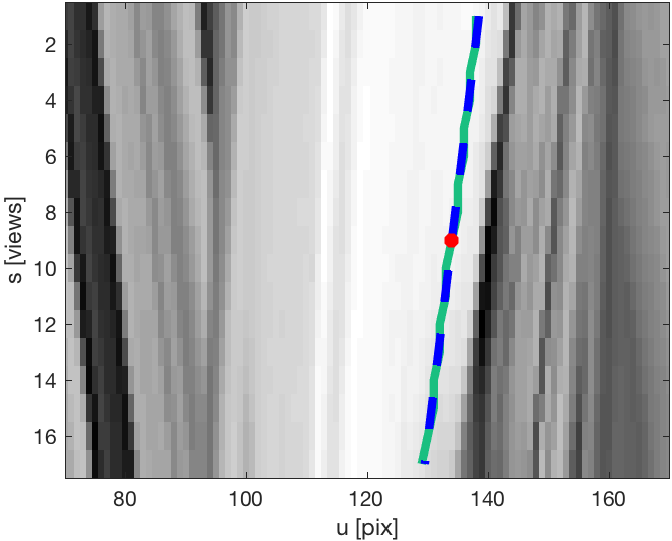}\label{fig:smallEpiHorz}}\hfil
\subfloat[][]{\includegraphics[width=0.49\columnwidth,height=0.37\columnwidth]{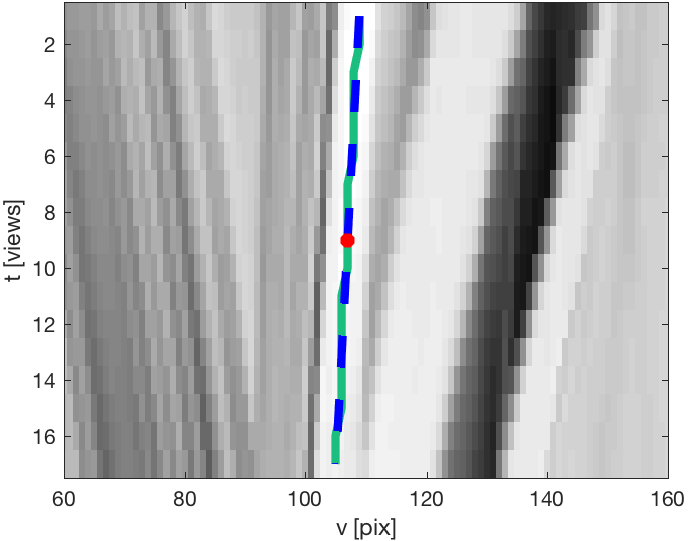}\label{fig:smallEpiVert}}
\caption{Sample (a) horizontal and (b) vertical EPIs from the crystal ball LF with small baseline for refracted feature (red). Extracted feature curves (green) match the plane of best fit (dashed blue). Refracted features appear almost linear, and are thus much more difficult to detect; however, slope consistency is still an important mechanism for distinguishing refraction.}
\label{fig:exEpiSmall}
% \vspace{-1.0em}
\end{figure}

Table~\ref{tbl:comparison} shows the results for baseline comparison, with the large baseline case shown at the top in Fig.~\ref{fig:comparison}.
For large baselines, a significant apparent motion was observed in the EPIs, and thus refracted features yielded non-linear curves, which strongly deviated from the hyperplanes in 4D. In contrast, the non-linear characteristics of refracted feature curves were much less apparent in shorter-baseline LFs.
Fig.~\ref{fig:exEpiSmall} shows the horizontal and vertical EPIs for a sample refracted feature taken from the small baseline LF. The feature curves appear almost straight, despite being refracted by the crystal ball. However, we observed that the slopes were inconsistent in this particular example, which could still be used to distinguish refracted features.

Our method's TPR was 34.3\% higher Xu's method for the large baseline case, which we attributed to more accurately fitting the plane, as opposed to the single hyperplane in 4D. For the small baseline case, we attributed our 38.9\% TPR increase to accounting for slope consistency, which Xu did not address.
Our slightly higher FPR would not be problematic as long as there are sufficiently many features. These false positives were due to occlusions, which are not yet distinguished in our implementation. However, this may still be beneficial as occlusions are non-Lambertian, and thus undesirable for most algorithms.
% show: large Xu (good), large us (good), small Xu (bad), small us (good).
% Also recall that we are only using the cross of the LF, which is a minimal sampling of the LF for plane-fitting.
Sampling from all the views in the LF would likely improve the results for both Xu's and our methods, as more data would improve the planar fit.

\begin{comment}
Crystal Large:
# features = 264
# refractive features in central view 35
# Xu's detected refractive features: 13
# Xu's false positives: 3
# Our detected refractive features: 25
# Our false positives: 8

Crystal Small:
# features = 197
# refractive features: 18
# Xu's refractive features: 3
# Xu's false positive: 0
# our refractive features: 10
# our false positive: 7
\end{comment}

\begin{ctable}[%
 caption = {Comparison for baselines and refractive objects in distinguishing refracted features$~^\textit{a}$},
 label = tbl:comparison,
 pos = t!
 ]
 {c cc cc}{
\tnote[a]{For detection, true positive rate (TPR), higher is better, and false positive rate (FPR), lower is better. Over 10 LFs for each case.\vspace{-4.0em}}
 }
 { \FL  & \multicolumn{2}{c}{\textbf{Xu's}} & \multicolumn{2}{c}{\textbf{Proposed}} \\
  & \textbf{TPR} & \textbf{FPR} & \textbf{TPR} & \textbf{FPR} \ML
  \textbf{Baseline} & & & & \\
  large (275 mm) & 37.1\% & \textbf{1.3\%} & \textbf{71.4\%} & 3.5\% \\
  small (64 mm) & 16.6\% & \textbf{0\%} & \textbf{55.5\%} & 3.9\% \\
 \FL \textbf{Object} & & & & \\
  cylinder & 12.7\% & 14.8\% & \textbf{90.9\%} & \textbf{10.1\%} \\
  sphere & 27.4\% & 54.2\% & \textbf{53.4\%} & \textbf{6.2\%} \LL
 }
\end{ctable}
%%%%%%%%%%%%%%%%%%%%%%%%%%%%%%%%%%%%%%%%%%%%%%%%%%%%%%%%%%%%%%%%%%%%%%%%%%%%%%%%
\subsection{Detection with the Lenslet-based LF Camera}

% Uncalibrated LFs were used due to time constraints; calibration would likely improve the results as systematic lens distortions cause Lambertian features to appear nonlinear at the extreme views.

We investigated two different types of refractive objects, including a glass sphere and an acrylic cylinder, shown in the bottom two rows of Fig.~\ref{fig:comparison}. The sphere exhibited significant refraction along both the horizontal and vertical viewing axes; however, the cylinder only exhibited significant refraction perpendicular to its longitudinal axis.
The refractive objects were placed within a textured scene in order to create textural details on the refractive surface.

Table~\ref{tbl:comparison} shows the results of over 10 LFs taken from a variety of different viewing poses of the given refractive object.
% As with the previous experiments in Section A, Xu's method was unable to differentiate refractive features with the smaller baseline of the plenoptic camera. For both cases, our method succeeded with a much higher DR because of slope consistency.
Planar error did not appear to be a strong indicator of refraction for the lenslet-based LF camera. Fig~\ref{fig:comparison} shows a sample detection for the cylinder. Xu's method was unable to detect the refractive cylinder, while our method succeeded.
Since the cylinder was aligned with the vertical axis, we expected non-linear behaviour for cylinder features along the horizontal axis. However, the small baseline of the camera reduced this effect, yielding EPIs similar to Fig.~\ref{fig:exEpiSmall}, and so the warping due to refraction through the cylinder was not apparent from this measure alone.
% The large planar errors on the sides of the image are due to lens distortion and insufficient thresholds, which may be corrected through better ridge-following in the correlation EPI, and further calibration of the LF camera.
On the other hand, slope consistency was a very strong indicator of refraction. % \dorian{which we can illustrate in PR-curves vs. slope consistency threshold vs planar error threshold.}
% Fig.~\ref{fig:slopeDiff} shows the features in the LF central view with the slope consistency depicted in colour, where warmer colours indicate larger inconsistency. The cylinder is clearly indicated, with more consistency towards the edges of the cylinder as features are wrapped around the cylinder's surface and begin to appear more linear.
% \dorian{Can we infer refractive shape from the distribution of slope differences?}

A refractive sphere was also investigated.
% , and exhibited more refractive properties than the cylinder. Fig.~\ref{fig:sphereEpi} shows the slightly curved lines from a vertical feature's EPI. Interestingly, the white line at $v=200$ pix is a specular reflection, and clearly appears as a super-position in the EPI, as Wanner discussed in~\cite{wanner2013reflectiveTransparentFromEPI}.
A comparison of Xu's and our method is shown in Fig.~\ref{fig:comparison}, whereby our method successfully detected the refracted features, while Xu's failed to reliably detect the sphere.
% how can we explain the figure? Some
Features that were located close to the edge of the sphere appear more linear, and thus were not always detected. Other missed detections were due to specular reflections, that appeared like well-behaved Lambertian points. Finally, there are some missed detections near the middle of the sphere, where there is identical apparent motion in the horizontal and vertical hyperplanes.
\subsection{Rejecting Refracted Features for Structure from Motion}

We validated our method by examining the impact of refracted features on an SfM pipeline. We captured a sequence of LFs that gradually approached a refractive object using a lenslet-based LF camera; thus the image sequence had an increasing number of refracted features.
We used Colmap, a publicly-available, modular SfM implementation~\cite{schoenberger2016sfmRevisited}. The centre view of the LF was used as input to SfM. Incremental SfM was performed on an image sequence where each successive image had an increasing number of refracted features, making it increasingly difficult for SfM to converge.
If SfM converged, a sparse reconstruction was produced, and the reprojection error was computed.

For each LF, SIFT features in the central view were detected, creating an unfiltered list of features, some of which were refractive. Our distinguisher was then used to remove refracted features, creating a filtered list of features. Both the unfiltered, and filtered lists of features were imported separately into the SfM pipeline, which included its own outlier rejection and bundle adjustment.

Outlier rejection schemes, such as RANSAC, are often used to reject inconsistent features, which includes refracted features. And while we observed some sequences where RANSAC successfully rejected most of the refracted features, more than 53\% of inliers were actually refracted features in some cases. This suggested that in the presence of refractive objects, RANSAC is insufficient on its own for robust and accurate structure and motion estimation.

\begin{figure}[tbp]
\centering
\subfloat[][]{\includegraphics[width=0.95\columnwidth]{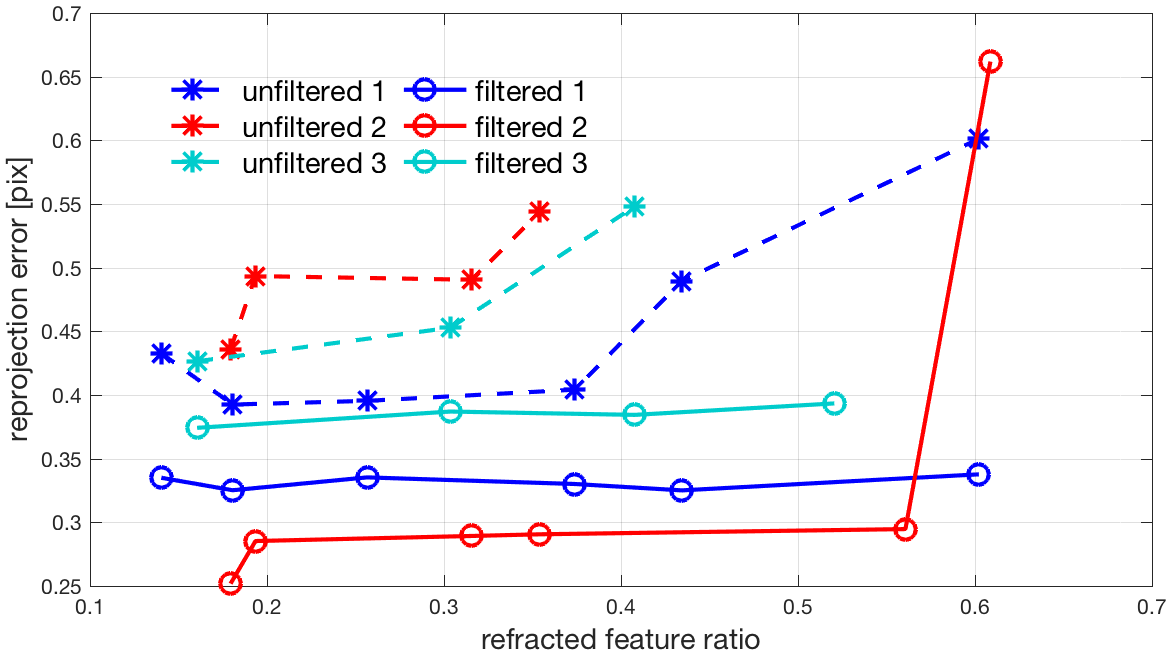}\label{fig:sfmResults}}\\
\subfloat[][]{\includegraphics[width=0.48\columnwidth]{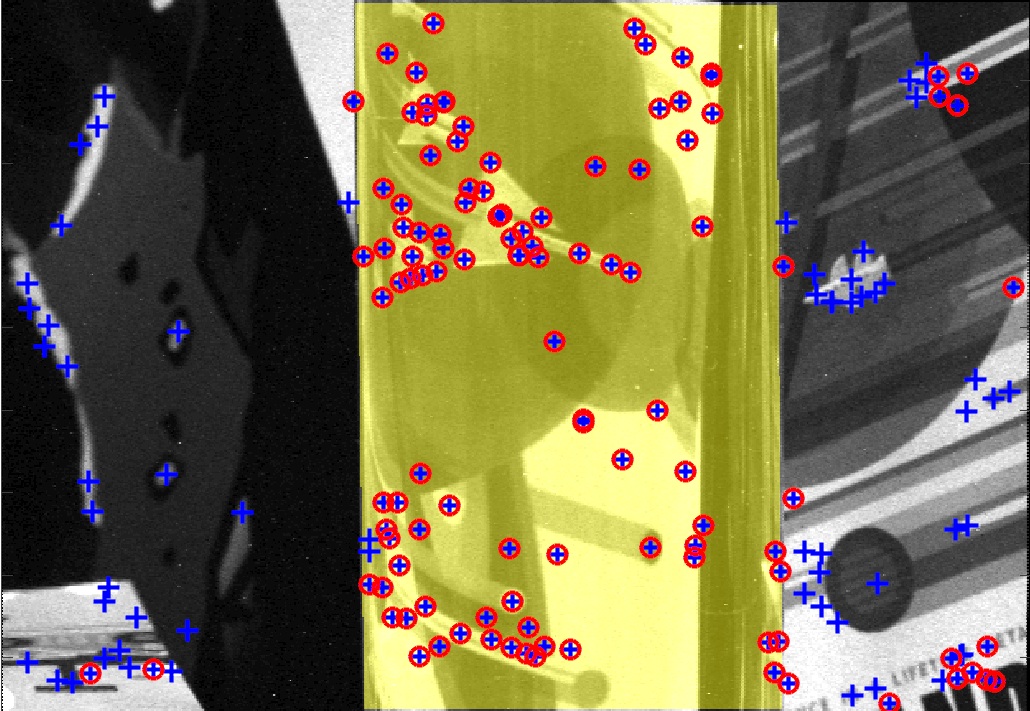}\label{fig:cylinderFail}}\hfil
\subfloat[][]{\includegraphics[width=0.48\columnwidth]{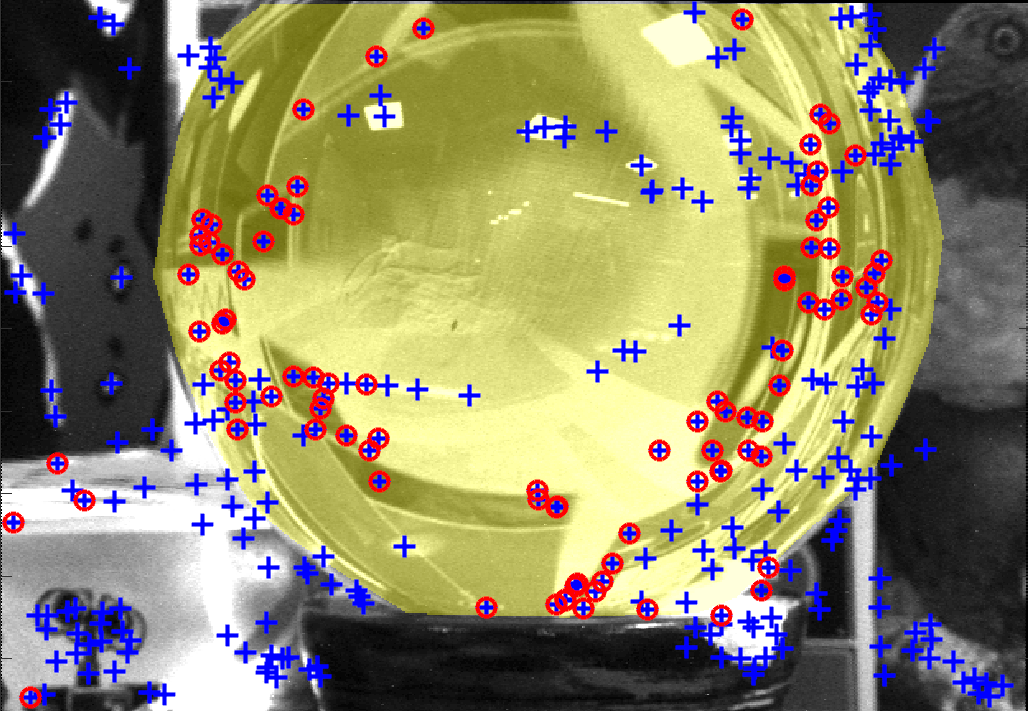}\label{fig:sphereFail}}
\caption{(a) Rejecting refracted features with our method yielded lower reprojection errors and better convergence.  SfM reprojection error vs refracted feature ratio for the unfiltered case containing refracted features (blue), and filtered case (red).
% The filtered cases were lower in error and , suggesting better performance. % For sequences 2 and 3, the unfiltered cases failed after the 0.4 refracted feature ratio, while the corresponding filtered cases converged for much longer, suggesting better convergent behaviour.
The spike in error at 0.6 ratio for filtered sequence 2 is due to insufficient inliers after RANSAC.
(b) and (c) show example images for the refractive cylinder and sphere (shown in yellow), respectively, that could not converge without pre-filtering for refracted features using our method. Detected features are shown in blue crosses, with features identified as refractive shown in red circles.}
\label{fig:sfmFail}
\vspace{-1.0em}
\end{figure}

We measured the ratio of refracted features $r = i_{r} / i_{t}$, where $i_{r}$ is the number of refracted features in the image (obtained via a manually-defined mask), and $i_{t}$ is the total number of features detected in the image. We considered the reprojection error as it varied with $r$. 
The results are shown in Fig.~\ref{fig:sfmResults}, in which, unsurprisingly, the error for the unfiltered case was consistently significantly higher (up to 42.4\% higher for $r < 0.6$ in the red case), and increased much faster than the filtered case, except when the number of inlier features became too low ($<30$). This suggested that having a more consistent (non-refractive) feature set improves the accuracy of reconstruction. Additionally, in many cases the unfiltered case failed to converge, while the filtered case was still successful, suggesting better convergent properties. Sample scenes that prevented  SfM from converging are shown in Fig.~\ref{fig:cylinderFail}, and~\ref{fig:sphereFail}. These scenes could not be used with SfM without our refracted feature distinguisher.

However, rejecting refracted features prior SfM does not guarantee convergence for SfM, which can fail for a number of reasons. Perhaps most significantly, removing all of the refracted features reduces the number of candidate features for matching, sometimes below the threshold for minimum number of features required by Colmap to register images. In these situations, the unfiltered case is sometimes more likely to converge (albeit with much larger error) simply due to having more inlier features. This is seen in filtered case 2 in Fig.~\ref{fig:sfmResults}. Additionally, as we move closer, the number of detected features naturally decreases, making SfM in the presence of refractive objects even more challenging.

% when we run SfM normally, we get phantom points, which are X. By applying our method, we don't get them. This is important because...

For the cases where SfM was able to converge in the presence of refractive objects, we observed ``phantom points'' in the SfM reconstruction. Phantom points are 
% whose 3D positions did not correspond to any 3D object
points that were placed in empty space near---but not on---the refractive object by SfM, due to refracted features counted as inliers.
%we observed that the unfiltered case often produced ``phantom'' points, where refracted features were incorrectly modelled when counted as inliers. %This is illustrated in Fig.~\ref{fig:sfmStructCyl10}, with the refracted features outlined in red for the unfiltered case.
With our method, there were little to no such phantom points in the reconstruction. This is a subtle but important difference since the absence of information is treated very differently from incorrect information in robotics. For example, phantom points might incorrectly fill an occupancy map, preventing a robot from grasping refractive objects.

\section{CONCLUSIONS}

In this paper, we proposed a method to distinguish refracted features based on a planar fit in 4D and slope consistency. To achieve this, we extracted feature curves from the 4D LF using textural cross-correlation. For large baselines, our approach yielded higher rates of detection than previous work; however, for smaller baselines, including a lenslet-based LF camera, previous methods were unable to detect refracted objects, while our approach was successful. For these baselines, slope inconsistency proved to be a much stronger indicator of refraction than planar consistency.
This is appealing for mobile robot applications, such as domestic robots that are limited in size and mass, but will have to navigate and eventually interact with refractive objects.
% Our refractivity distinguisher demonstrated higher detection rates than previous works, and can be applied to LF plenoptic cameras with much smaller baselines than before, which is appealing for mobile robot applications.

We also demonstrated that rejecting refracted features in monocular SfM yields % 
lower reprojection errors, which may imply better reconstructions of the non-refractive parts of the scene in the presence of refractive objects. 
% We provided sample scenes that SfM could not handle without our method.
%, and we do not create incorrect structure of the refracted object. 
Further research into slope consistency, the distribution of refracted features, and LF-specific features may lead towards recovering refractive shape from features.

% Takeaway from paper. Who this might be useful for. Future work.
It is important to note that while we have developed a set of criteria for refracted features in the LF, these criteria are not necessarily limited to refracted features. Depending on the surface, specular reflections may appear as non-linear. Poor camera calibrations % (which are technically due to refraction) 
may also cause Lambertian features to appear refractive in the light field. Occlusions are also occasionally detected, though they must be properly identified in future work. These types of features are typically undesirable, and so we retain features that are strongly Lambertian, and thus good candidates for matching, which ultimately leads to more robust robot performance in the presence of refractive objects.
% Thierry: we throw away all the shitty kids on the block and keep the good ones

Finally, in this paper, we explored the effect of removing the refractive content from the scene. In future work, we plan to exploit the refractive content for robot motion and refractive shape recovery.
%%%%%%%%%%%%%%%%%%%%%%%%%%%%%%%%%%%%%%%%%%%%%%%%%%%%%%%%%%%%%%%%%%%%%%%%%%%%%%%%
\addtolength{\textheight}{-11cm}   % This command serves to balance the column lengths

%%%%%%%%%%%%%%%%%%%%%%%%%%%%%%%%%%%%%%%%%%%%%%%%%%%%%%%%%%%%%%%%%%%%%%%%%%%%%%%%

%%%%%%%%%%%%%%%%%%%%%%%%%%%%%%%%%%%%%%%%%%%%%%%%%%%%%%%%%%%%%%%%%%%%%%%%%%%%%%%%

%%%%%%%%%%%%%%%%%%%%%%%%%%%%%%%%%%%%%%%%%%%%%%%%%%%%%%%%%%%%%%%%%%%%%%%%%%%%%%%%
\bibliographystyle{IEEEtran}
\bibliography{IEEEabrv,DorianTsaiReferencesBib}

\begin{comment}
\dorian{TODO:
\begin{itemize}
	\item cut down to 8 pages
	\item ``detection rate'' to ``distinction/distinguishing rate'' or ``true positive rate''
	\item true positives, false negatives, false positives, true negatives, what is ``good'' for this binary classifier, under what conditions, consider subsubsection
	\item make feature markers larger
	\item PR curves wrt slope consistency vs planar thresholds
	\item camera/arm pose experiments - quantitative SfM results
\end{itemize}
}
\end{comment}
\end{document}